%% file: iclr2027_conference.tex
\documentclass{article}
\usepackage{iclr2027_conference,times}
\usepackage{graphicx,xcolor}
\usepackage{booktabs}
\usepackage{hyperref}
\usepackage{url}
\usepackage{wrapfig}
\usepackage{caption,subcaption}
\usepackage{placeins}
\usepackage{multirow,makecell}

\input{math_commands}

\title{De-Diffusing WAMs: Action-Sufficient Direct Prediction for Robot Control}
\author{\bfseries Shaoyi Zheng$^{1,2}$,\quad Zhongtao Jiang$^{1}$,\quad Yue Peng$^{2}$,\quad Wenbo Lu$^{2}$, \\[2pt]
\bfseries Shunchi Zhang$^{1}$,\quad Xi Wang$^{2}$,\quad Shengji Wan$^{2,\dagger}$}
\makeatletter
\renewcommand{\@maketitle}{\vbox{\hsize\textwidth
  {\LARGE\scshape\@title\par}
  \vskip 12pt
  {\centering
    \begin{tabular}{@{}c@{}}\@author\end{tabular}\par}
  \vskip 10pt
}}
\makeatother
\iclrfinalcopy

\begin{document}
\maketitle
\fancyhead{}
\renewcommand{\headrulewidth}{0pt}
\begingroup
\renewcommand{\thefootnote}{}
\footnotetext{\textsuperscript{1}ByteDance. \textsuperscript{2}New York University (NYU). \textsuperscript{\ensuremath{\dagger}}Corresponding author: Shengji Wan (\href{mailto:sw5973@nyu.edu}{\texttt{sw5973@nyu.edu}}).}
\endgroup

\begin{abstract}
Cascade world-action models (WAMs) use predicted visual futures to condition robot actions, but inherit costly iterative inference from diffusion-pretrained experts.
We ask what a single-step predicted future must preserve to support accurate single-step action prediction.
In the evaluated WAM, resampling video noise has little effect on downstream actions, yet naively truncating both streams to one step substantially reduces task success.
Video-space supervision only partially closes this gap, and a local sensitivity analysis shows that vision prediction errors have highly unequal consequences for action accuracy.
These findings motivate \textbf{DirectWAM}, which converts pretrained video and action experts into single-step predictors without stochastic sampling through \emph{action-sufficient distillation}, while retaining the explicit future-to-action cascade.
After learning a one-step action model, we train the future predictor by decoding actions from its own predicted futures and backpropagating demonstration action error through the expert, with an optional future-latent reconstruction loss.
The resulting objective evaluates the student-generated future through its downstream control consequences rather than relying on representation matching alone.
DirectWAM achieves 97.1\% average success across four LIBERO suites and 85.4\% on RoboTwin 2.0, outperforming Flash-WAM at the same one-evaluation-per-stream budget. 
This advantage has also extended to three real-world tasks that we designed.
Compared with the full iterative teacher, it reduces inference latency by $28.0\times$.
\end{abstract}

\section{Introduction}
\label{sec:introduction}

Cascade world-action models (WAMs) leverage pretrained video models to predict visual futures, which then condition an inverse dynamics model (IDM) for action generation.
Unlike standalone video generation, the predicted future is not the final output but an intermediate representation for robot control.
This causal future-to-action relationship allows a robot policy to utilize world prior from video while retaining futures that can be decoded for inspection.
However, diffusion-based video and action experts realize this computation through repeated denoising, introducing a substantial inference cost.
We study how to retain the pretrained experts and explicit future-to-action interface while replacing their iterative inference procedure with direct prediction.

\begin{figure}[t!]
    \centering
    \includegraphics[width=0.9\textwidth]{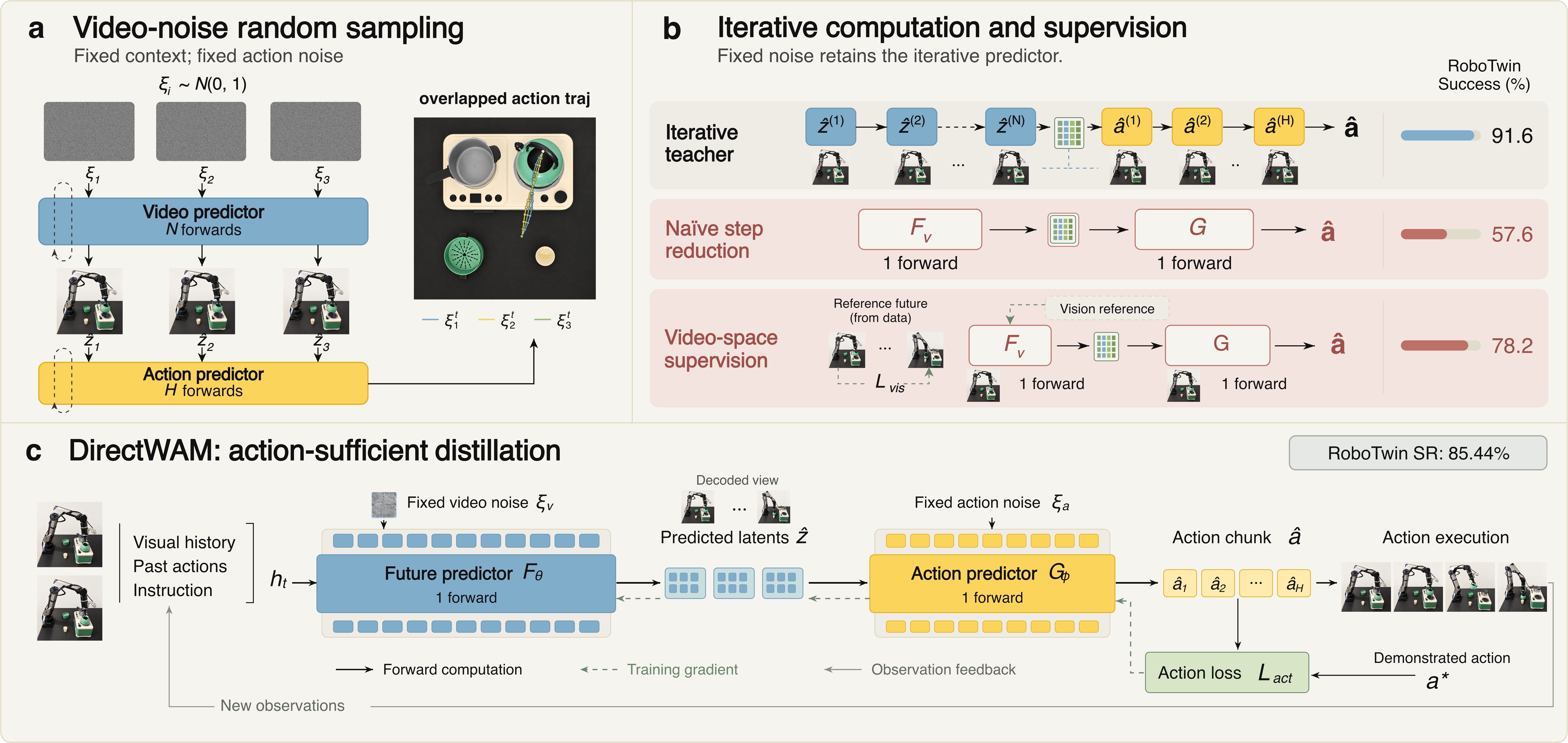}
    \caption{
        \textbf{Overview of DirectWAM.}
        (a) Video randomness produces similar downstream actions.
        (b) Naive step truncation and visual supervision induce a noticeable performance gap.
        (c) DirectWAM is a deterministic one-step cascade WAM trained through action-sufficient distillation.
    }
    \label{fig:main}
\end{figure}

Recent acceleration methods either modify the future-to-action interface or compress its computation.
Fast-WAM retains video co-training but removes explicit future prediction at deployment~\cite{fastwam}; Motus supports action-only inference alongside other joint and conditional modeling modes~\cite{motus}.
Among methods that retain the cascade, Flash-WAM~\cite{flashwam} uses modality-aware consistency distillation to achieve one-step inference in both the video and action streams.
DIDO~\cite{dido} concentrates supervision on control-relevant visual regions to learn single-step future prediction, while retaining iterative action generation.
These advances establish that cascade inference can be accelerated.
However, a more basic question is largely overlooked in WAM acceleration: what must a single-step future preserve to support accurate downstream action prediction?
As the future serves only as an intermediate representation, the reconstruction objective of video diffusion may not naturally align with the control objective of the downstream IDM.

We first disentangle the roles of stochastic sampling and iterative refinement, using LingBot-VA as our experimental instantiation.
Varying only the video noise produces limited changes in downstream actions, and fixing it throughout an episode barely affects closed-loop success.
Stochasticity can therefore be removed, yielding a deterministic future predictor that nonetheless still relies on the teacher's multi-step computation.
Removing this iteration is non-trivial: naively reducing the steps degrades control, and learning a one-step predictor with video-space supervision recovers only part of the gap.
Therefore, these results motivate learning the direct future predictor through the action model that consumes its predictions, rather than relying on video-space supervision alone.

To understand how the remaining prediction errors affect control, we selectively correct components of the future-representation error under a frozen action expert.
Correcting only a small, locally action-sensitive component recovers a disproportionate share of the achievable offline action-error reduction.
The complementary component still contributes, but the imbalance shows that representation-error magnitude is a poor proxy for downstream importance.
This motivates \emph{action sufficiency} as the learning principle for single-step prediction: the predicted future should be optimized according to the action decoded from it, rather than through representation matching alone.

We implement this principle in \textbf{DirectWAM} through \emph{action-sufficient distillation}.
Starting from diffusion-pretrained video and action experts, we first train a single-step action expert with the video expert fixed.
We then train the single-step future predictor end-to-end through this action expert: actions are decoded from the student's own predicted futures, and the demonstration action error is backpropagated into the future predictor.
An optional future-latent reconstruction loss anchors predictions to the demonstrated future.
Because both objectives supervise final predictions rather than intermediate denoising states, the pretrained experts are converted into a direct cascade without the student having to reproduce the teacher's denoising trajectory.

In experiments, DirectWAM achieves 97.1\% average success on LIBERO and 85.4\% on RoboTwin 2.0, outperforming cascade baselines at the same compute budget and substantially narrowing the single-step control gap. These gains further carry over to three real-world manipulation tasks.
Inference latency decreases from 5042\,ms for the full iterative teacher to 180\,ms for DirectWAM, a $28.0\times$ speedup.
Together, these results demonstrate a practical route to retaining explicit future prediction while replacing iterative diffusion inference with direct computation trained for downstream control. Our contributions are threefold:
\begingroup
\setlength{\leftmargini}{1.5em}
\setlength{\topsep}{3pt}
\setlength{\partopsep}{0pt}
\begin{enumerate}
\setlength{\itemsep}{2pt}
\setlength{\parsep}{0pt}
    \item \textbf{Separating stochastic sampling from iterative refinement.}
    We characterize their distinct roles in the evaluated cascade WAM: video-noise resampling has limited downstream effects, whereas naively truncating both streams to one step causes control drops.

    \item \textbf{Action sufficiency as a learning criterion.}
    We show that the magnitude of future-prediction error is a poor proxy for its effect on action accuracy, motivating a training objective that supervises the student-generated future through the actions it induces online.
    
    \item \textbf{A deterministic single-step future-to-action cascade.}
    We introduce \textbf{DirectWAM}, a one-step de-diffused cascade WAM that preserves explicit future prediction while achieving strong control performance both in simulation and in the real world.
\end{enumerate}
\endgroup

\section{Related Work}
\label{sec:related_work}

\paragraph{World-action models and future-conditioned control.}
Video-based control uses predicted observations as an intermediate representation for action generation~\citep{unipi}.
Recent World-Action Models (WAMs) more tightly integrate visual dynamics and action prediction: LingBot-VA through autoregressive video--action modeling~\cite{lingbotva}, DreamZero with a video-pretrained backbone~\cite{dreamzero}, and Motus within a multi-expert architecture~\cite{motus}.
Building on the cascade WAM interface, we investigate how to learn direct future prediction according to its action consequences, rather than assuming that control requires fully refined video.

\paragraph{Distillation for efficient cascade WAMs.}
Consistency models and distribution matching distillation provide general
routes to reducing diffusion sampling steps~\cite{consistency_models,dmd}.
A distinct architectural approach, Fast-WAM, retains video co-training but
removes explicit future prediction at deployment~\cite{fastwam}.
We instead focus on acceleration that preserves the future-to-action cascade.
Flash-WAM addresses the different noise regimes of video and action streams
by assigning modality-specific consistency functions, enabling single-step
inference in both streams~\cite{flashwam}.
DIDO combines distribution matching with interaction-centric representation
guidance for one-step video prediction, using explicit interaction
representations together with visual feature alignment to preserve
manipulation-relevant information~\cite{dido}.
These methods demonstrate that efficient cascade WAM prediction requires
more than applying a generic sampling shortcut.
DirectWAM pursues the same goal of retaining useful future prediction, but
defines its learning criterion through the existing action model.

\paragraph{Control-oriented predictive representations.}
Predictive accuracy need not align with downstream decision quality, as studied through objective mismatch~\cite{objective_mismatch}, value-aware model learning~\cite{value_aware}, and value equivalence~\cite{value_equivalence}. Recent WAMs address related representation challenges through geometric or semantic supervision: MECo-WAM~\cite{mecowam} transfers action-relevant 4D geometric priors from a frozen VGGT~\cite{vggt}, while Robust-WAM augments video-generation-based policies with future-semantic alignment~\cite{robustwam}. GigaWorld-Policy-0.5~\cite{gigaworld} strengthens action–dynamics coupling through action-conditioned future modeling while allowing action-only inference. Our work shares the broader goal of learning predictive information useful for control, but focuses on how a predicted future should be learned when it must condition a single-step action predictor. Specifically, DirectWAM converts diffusion-pretrained experts into a deterministic single-step cascade while retaining their future-representation interface. Rather than introducing a new semantic or geometric target space, it trains the future predictor through demonstration-action error decoded from its own predictions by a learned direct action model, with optional future-latent anchoring. This places the learning criterion at the interface between the student-generated future and its action consumer, while preserving explicit future conditioning with one evaluation per stream.

\section{Rethinking Future Prediction for Cascade WAMs}
\label{sec:rethinking_diffusion}

In this section, we systematically study the effect of the future prediction in cascade WAMs on downstream robot control. The future is not the endpoint of the model: its effect on control is mediated by the action model and should therefore be viewed from a control perspective.

\subsection{DOES VIDEO-SIDE STOCHASTICITY MATTER FOR CONTROL?}
We begin by evaluating whether the action model is sensitive to the stochasticity of future prediction.
Diffusion provides the video stream with the ability to sample diverse futures under the same context, but it is unclear whether this variation is propagated into different actions.
We therefore ask: \emph{does resampling the imagined future produce meaningfully different control decisions?}

To isolate the effect of stochastic future sampling, we fixed the action-side noise across samples, while varying only the video-diffusion noise.
Thus, any variation in the generated actions  is induced by changing the sampled future,
rather than by resampling the action model itself.

\paragraph{Video-side sampling variation is strongly attenuated by the downstream action model.}
We first ask how much variation induced by changing only the video-diffusion noise survives after the action mapping.
For two samples from the same context, let $D_V$ denote their distance in future-representation space and $D_A$ the corresponding distance between their induced actions.
We evaluate variation propagation through three complementary statistics:
(i) the Pearson correlation between $D_V$ and $D_A$ across same-context sample pairs,
(ii) the mean scale ratio $D_A/D_V$, and
(iii) the fraction of total variance attributable to within-context sampling variation, computed separately in video and action space.
Experiment setting details are provided in Appendix~\ref{app:sampling_metrics}.

\begin{table}[htbp]
    \centering
    \caption{
    Three metrics measuring the effect of video variation on downstream actions. Same-context video-sampling variation is substantially attenuated
    by the action model.
    }
    \label{tab:sampling_propagation}
    \begin{tabular}{lcccc}
        \toprule
        Benchmark & Contexts $\times$ samples & $\rho(D_V,D_A)$ & Mean $D_A/D_V$ & V $\to$ A var ratio \\
        \midrule
        RoboTwin & $237 \times 32$ & 0.69 & 0.025 & $6.6\% \to 0.030\%$ \\
        LIBERO   & $64 \times 16$  & 0.80 & 0.138 & $1.6\% \to 0.25\%$ \\
        \bottomrule
    \end{tabular}
\end{table}

Table~\ref{tab:sampling_propagation} shows that video sampling variation is positively correlated with variation on the action side, but its magnitude is substantially attenuated.
On RoboTwin, the mean action-space distance is only $2.5\%$ of the corresponding video-space distance, and the relative contribution of sampling variance shrinks from $6.6\%$ in video space to $0.030\%$ in action space.
The same trend holds on LIBERO, although with weaker attenuation. 
Additional clustering and multimodality analyses supporting this conclusion are provided in Appendix~\ref{app:action_modes} and~\ref{app:multimodality_test}.

All the above tests support a consistent conclusion: measurable video-sampling variation rarely expands into clearly separated downstream control decisions.

\begin{figure}[!t] \centering \begin{minipage}{0.95\textwidth}
\captionsetup[subfigure]{skip=2pt} \begin{subfigure}[t]{0.32\linewidth} \vspace{0pt}\centering \raisebox{0pt}[120bp][0pt]{\includegraphics[width=\linewidth,height=120bp,keepaspectratio]{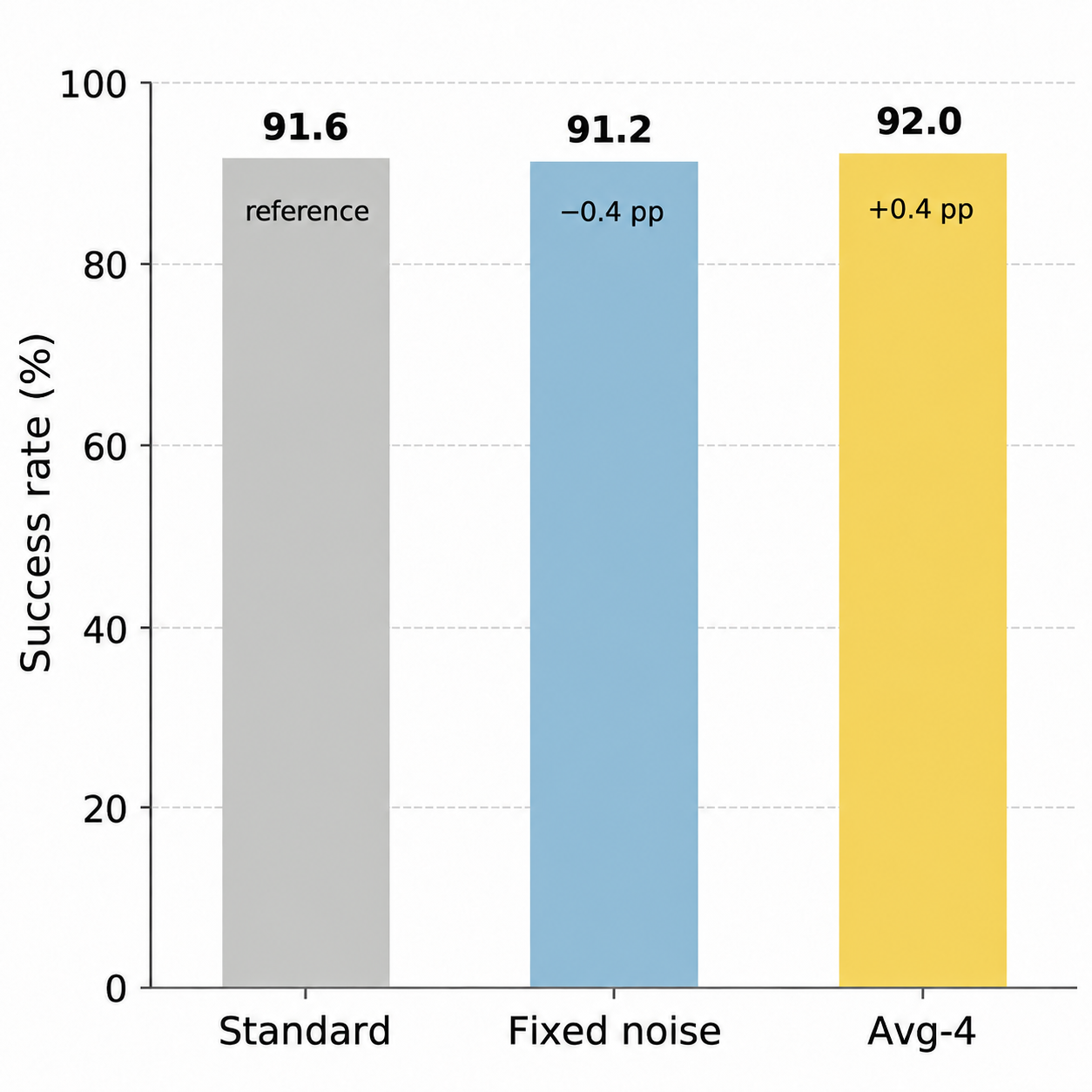}} \caption{}\label{fig:video_sampling_control}\label{tab:video_sampling_control} \end{subfigure}\hfill \begin{subfigure}[t]{0.64\linewidth} \vspace{0pt}\centering \raisebox{0pt}[120bp][0pt]{\includegraphics[width=\linewidth,height=120bp,keepaspectratio]{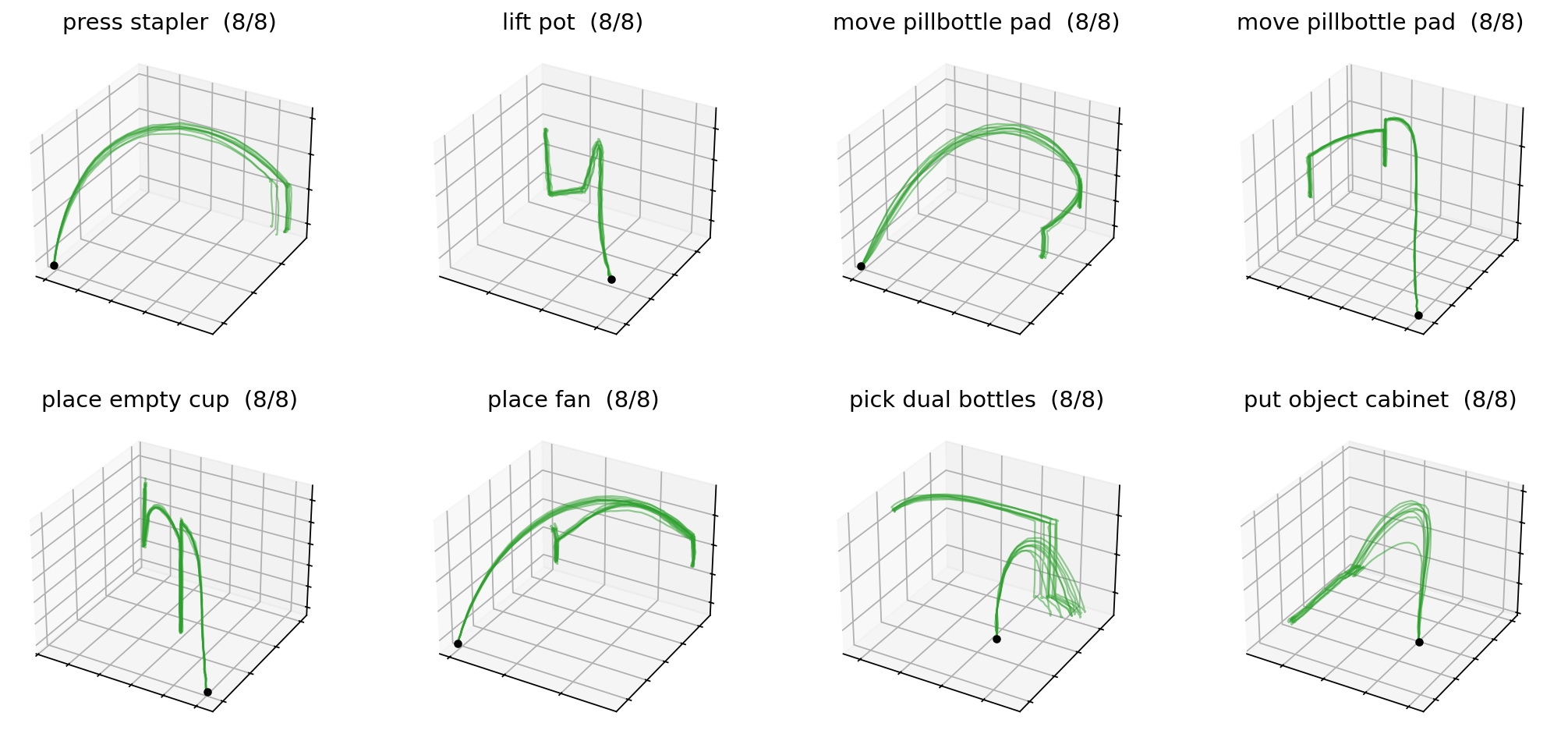}} \caption{}\label{fig:closedloop_traj} \end{subfigure}
\end{minipage}
\caption{\textbf{Closed-loop effects of video-side stochasticity.} \textbf{(a)} RoboTwin success rates of the teacher, fixed video noise, and averaging four future predictions. \textbf{(b)} Eight rollouts per scene under different video-noise seeds. Successful trajectories remain closely aligned.} \label{fig:video_stochasticity} 
\end{figure}

\paragraph{Different video samples rarely induce distinct closed-loop behaviors.}
Beyond offline metrics, we extend experiments to the closed loop tasks to test the effect of the future randomness in practice.

We intervene directly on video-side stochasticity with two complementary variants.
\emph{Fixed noise} reuses a single in-distribution video-noise realization
throughout the episode, $\xi_{v,t}=\bar\xi_v$ for all $t$.
\emph{Avg-4} averages four independently sampled future representations,
$\bar z_t=\operatorname{mean}(z_t^{(1)},\ldots,z_t^{(4)})$,
and passes the mean to the action model.\
The two interventions suppress different aspects of video-side stochasticity:
\emph{Fixed noise} suppresses temporal resampling across control chunks,
whereas \emph{Avg-4} suppresses sample-specific variation within each chunk by
averaging multiple futures.
As shown in Fig.~\ref{fig:video_sampling_control}, neither intervention causes
a systematic degradation in task success.


\begin{figure*}[!t]
    \centering
    \captionsetup[subfigure]{labelformat=parens,labelsep=none,skip=2pt}

    \begin{subfigure}[t]{0.6\textwidth}
        \vspace{0pt}
        \centering
        \includegraphics[
            width=\linewidth,
            height=1.15in,
            keepaspectratio
        ]{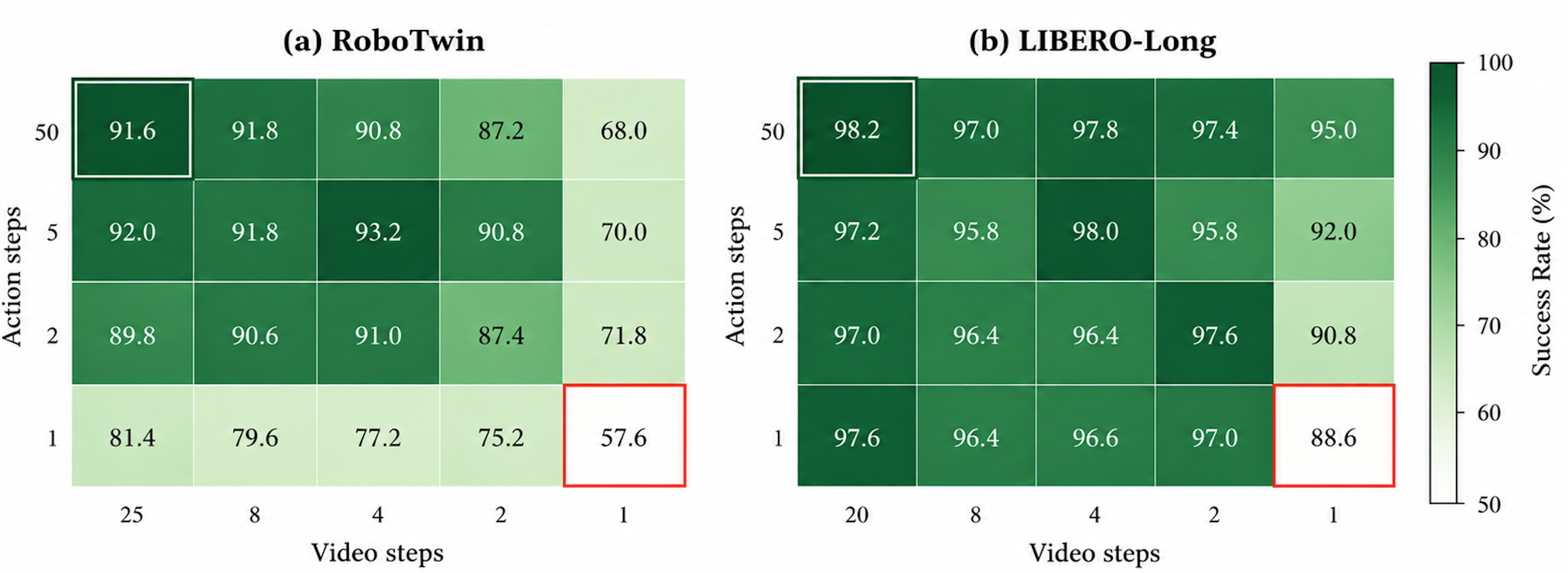}
        \caption{}
        \label{fig:inference_step_reduction}
    \end{subfigure}%
    \hfill
    \begin{subfigure}[t]{0.4\textwidth}
        \vspace{0pt}
        \centering
        \includegraphics[
            width=\linewidth,
            height=1.15in,
            keepaspectratio
        ]{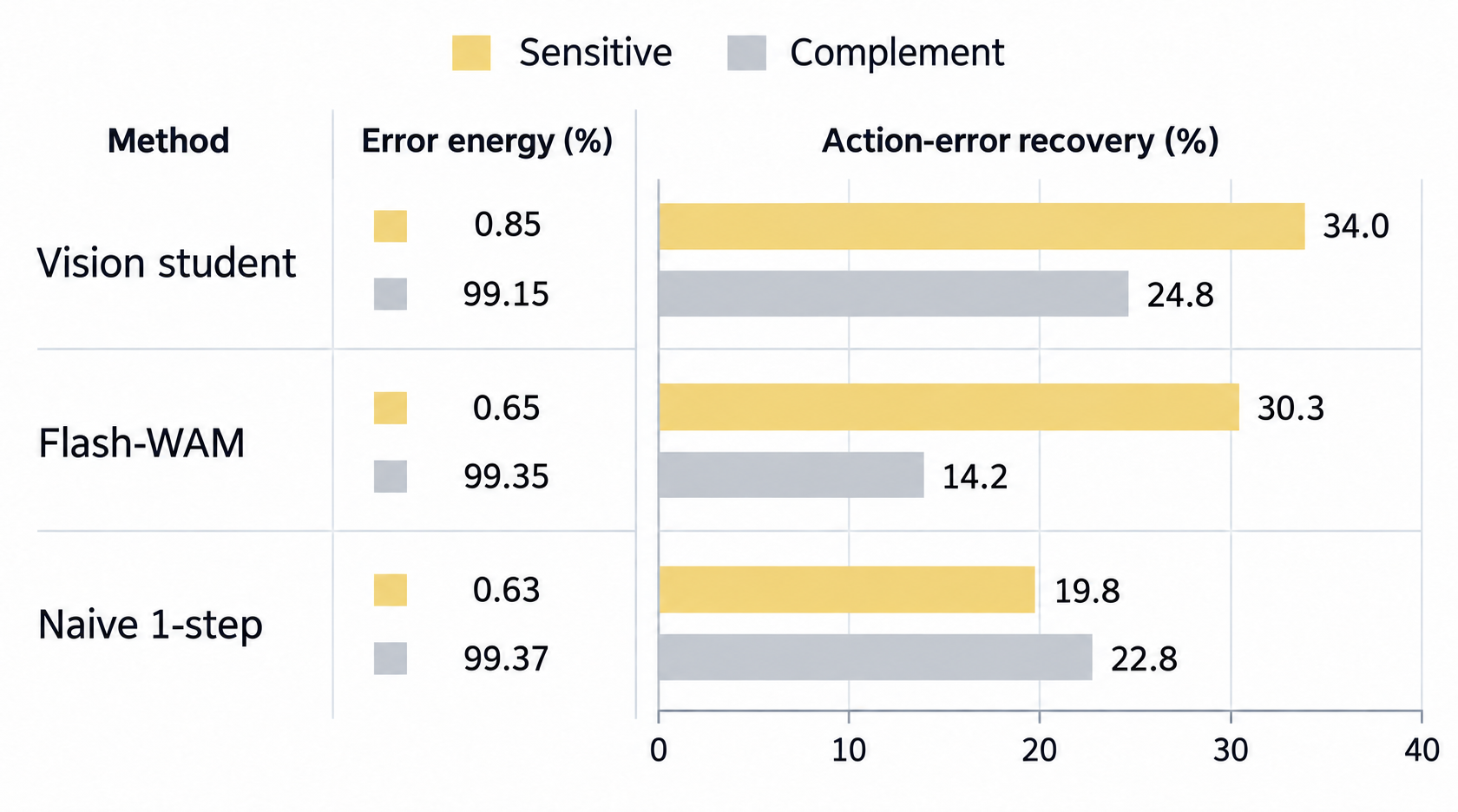}
        \caption{}
        \label{fig:action_sensitive_recovery}
        \label{tab:action_sensitive_replacement}
    \end{subfigure}

    \caption{
        \textbf{Inference-step reduction and action-sensitive error recovery.}
        \textbf{(a)} Success rates (\%) across video/action step budgets. 
        Performance degrades as the number of steps decreases.
        \textbf{(b)} The action-sensitive component holds only $0.63$--$0.85\%$ of the discrepancy energy yet recovers $19.8$--$34.0\%$ of the offline action-error improvement.
    }
    \label{fig:steps_and_recovery}
\end{figure*}

\paragraph{Limited Closed-Loop Trajectory Diversity.}
Success rates alone do not reveal whether video-noise sampling induces distinct execution strategies.
We therefore randomly sample the teacher eight times from the same initial state for each of 30 contexts selected for the strongest evidence of future branching, varying only the video-diffusion noise.
As shown in Fig.~\ref{fig:closedloop_traj}, successful rollouts follow closely aligned paths, with local deviations but no clear separation into distinct strategies.

We compute the RMS distance $\sigma(t)$ of the rollouts' end-effector positions from their mean position. We then average this spread over the time steps shared by all rollouts and divide by the episode's motion range $R$, giving $D_{\mathrm{traj}}=\operatorname{mean}_t[\sigma(t)]/R$.
The median normalized spread is $0.020$, accounting for a median absolute spread of $0.86\,\mathrm{cm}$ compared to a median motion range is $43\,\mathrm{cm}$.
Normalization details and outcome-specific summaries appear in Appendix~\ref{app:closed_loop_sampling}.
Video noise can alter outcomes in challenging contexts but induces limited closed-loop trajectory diversity.

\subsection{What Should a One-Step Future Preserve?}
\paragraph{One-step prediction requires more than truncation or video-space matching.}
The limited effect of video-noise sampling raises a separate question: how much iterative computation can be removed without retraining?
Figure~\ref{fig:inference_step_reduction} sweeps the video and action inference budgets on RoboTwin and LIBERO-Long.
Reducing both streams to two steps retains much of the teacher's performance: success changes from $91.6\%$ to $87.4\%$ on RoboTwin and from $98.2\%$ to $97.6\%$ on LIBERO-Long.
The transition to one step per stream is markedly more costly, lowering success to $57.6\%$ and $88.6\%$, respectively.
The streams differ in their sensitivity: RoboTwin degrades when either stream is reduced to one step, whereas LIBERO-Long largely tolerates action-step reduction until the video stream also reaches one step.
Thus, although some inference steps can be removed with little loss, the remaining refinement still contributes substantially to control.
Limited reliance on stochastic sampling does not imply that a useful one-step predictor is already available in the pretrained model.

\paragraph{Video-space matching remains insufficient.} We therefore ask whether the remaining gap can be closed by a learned one-step future predictor. To isolate this factor, all compared future predictors use the same model initialization, training data, optimization budget, and frozen single-step action expert; only the future-learning objective is changed. On RoboTwin, video-space supervision toward teacher-predicted or ground truth future representations achieves 76.0\% and 78.2\% success, respectively. While substantially stronger than naive one-step truncation, both remain well below the iterative teacher. Thus, improving representation-space fidelity alone does not fully recover the control performance of the original cascade. To understand this remaining gap, we next examine how future-representation errors translate into downstream action errors.

\paragraph{Representation-space fidelity is an incomplete proxy for control fidelity.}
We examine whether representation-space error is a reliable proxy for downstream action error.
Using the local Jacobian of a frozen action model, we decompose the difference between a predicted future $z^{\mathrm B}$ and the full-teacher future $z^{\mathrm T}$ into a locally action-sensitive component and its orthogonal complement:
\begin{equation}
\Delta z=z^{\mathrm T}-z^{\mathrm B},\qquad
\Delta z_{\mathrm{sens}}=P\Delta z,\qquad
\Delta z_{\mathrm{comp}}=(I-P)\Delta z.
\label{eq:error_decomposition}
\end{equation}
Here, $P$ is an orthogonal projection constructed from 32 sampled action-output directions.
We add each component separately to $z^{\mathrm B}$ and measure the resulting reduction in action error against demonstrations.
Recovery is normalized by the improvement obtained from replacing $z^{\mathrm B}$ with the full teacher prediction $z^{\mathrm T}$.
Appendix~\ref{app:teacher_sensitive_recovery} provides the projection and evaluation details.

Figure~\ref{fig:action_sensitive_recovery} shows that, across three variants, small components of the teacher-referenced representation discrepancy can have disproportionately large effects on the decoded action. Thus, representation distance alone does not indicate which prediction errors are most consequential for action prediction. A representation-matching objective therefore treats errors according to their size in representation space, rather than their effect on the downstream action model. This motivates using downstream action error as a direct learning signal for the one-step future predictor.

\subsection{DirectWAM: Action-Sufficient Distillation}
\label{sec:action_sufficient_distillation}
We introduce \textbf{action-sufficient distillation}, a model-conversion principle that defines the adequacy of a single-step predicted future of cascade WAMs by the action it enables, rather than by how faithfully it reproduces the teacher's iterative generative trajectory or full future visual representation.
Building on this principle, \textbf{DirectWAM} converts the diffusion-pretrained video and action experts of a cascade WAM into a de-diffused deterministic single-step future-to-action pipeline; we instantiate it with LingBot-VA in our experiments.

\paragraph{Learning the single-step action model.}
We first train the action model  $G_\psi$ to predict action chunks in a single evaluation. During this stage, the video pathway is frozen and the action model is conditioned on the demonstrated future representation, following the teacher-forced training interface of LingBot-VA. Supervision is applied directly to the final action prediction rather than to intermediate denoising states. This yields a single-step action readout, which is frozen in the subsequent future-learning stage and used to map each student-predicted future to its downstream action.

\paragraph{Action-sufficient future learning.}

The analysis above suggests that representation matching alone does not indicate which future-prediction errors matter most for downstream control. We therefore change both the training interface and the learning criterion of the cascade. The future predictor is evaluated on its own prediction through the single-step action model that consumes it, and the resulting action error serves as the primary supervision.
Given a conditioning context $h_t$ at time $t$, the future predictor $F_\theta$ produces a future representation $F_\theta(h_t)$. The action model $G_\psi$ then predicts an action chunk conditioned on this representation and $h_t$.
Let $\mathcal D$ denote the demonstration dataset and
$a^\star_{t:t+K-1}$ the target action chunk over a horizon of $K$ steps.
We minimize
\begin{equation}
\mathcal L_{\mathrm{ASD}}(\theta,\psi)
=
\mathbb E_{\mathcal D}
\left[
\ell_{\mathrm{act}}
\left(
G_\psi\!\left(F_\theta(h_t),h_t\right),
a^\star_{t:t+K-1}
\right)
\right],
\label{eq:action_sufficient_distillation}
\end{equation}
where $\ell_{\mathrm{act}}$ is the mean squared error loss.
The future predictor receives supervision through how the action
model translates its predictions into actions.
A predicted future satisfies this objective if it yields the
demonstrated action, even when it differs from the reference future.
This is the sense in which the future needs to be
\emph{action-sufficient}: its adequacy is determined by the action
it enables.

This differs from the original LingBot-VA training in two respects. First, LingBot-VA trains the action model under teacher forcing, conditioning it on ground-truth visual transitions, so its action loss does not supervise the consequences of errors made by the video model itself. Second, future prediction is trained as a generative objective in its own right. In DirectWAM, the student-predicted future instead passes through the deployment-time action interface, and downstream action error becomes the primary criterion for learning that future, while representation reconstruction is optional.

\paragraph{Optional Future Anchoring and Loss Composition.}
We optionally add a future-reconstruction anchor,
$\mathcal L_{\mathrm{anchor}}=\mathbb E_{\mathcal D}
[\|F_\theta(h_t)-z^\star_{t+1:t+K}\|_2^2]$,
where $z^\star_{t+1:t+K}$ is the demonstrated future latent chunk.
The overall objective is
\begin{equation}
\mathcal L_{\mathrm{train}}
=
\mathcal L_{\mathrm{ASD}}
+
\lambda_v \mathcal L_{\mathrm{anchor}}.
\label{eq:directwam_training_loss}
\end{equation}
Here, $\lambda_v$ controls the anchor strength;
$\lambda_v=0$ gives purely action-supervised future learning. The reconstruction term acts only as an optional regularizer that preserves proximity to the demonstrated future; downstream action error remains the primary criterion for learning the future representation. Training details and loss curves are provided in
Appendix~\ref{app:training_details}
and~\ref{app:stage2_dynamics}.

\paragraph{Direct Prediction WAM targeted at Robot Control}
In a nutshell, DirectWAM converts diffusion-pretrained video and action experts into a deterministic single-step cascade WAM, retaining the explicit future-to-action interface. The predicted future is learned primarily through the downstream action it induces under the principle of Action Sufficiency, with future reconstruction used only as an optional anchor. In this way, DirectWAM preserves the predictive prior of the pretrained WAM without retaining its stochastic iterative generative inference procedure.

\section{Experiments}
\label{sec:experiments}

\subsection{Hardware Setting}
\label{sec:hardware_setting}

We use the same GPU type for training and inference timing. Each GPU has 80 GB of memory, with nominal peak throughput of 67 TFLOP/s for FP32 arithmetic and approximately 990 TFLOP/s for dense BF16 matrix operations. We measure inference latency on a single GPU and report the median across 500 timed runs following 500 warm-up runs.

\subsection{Simulation Results}
\label{sec:simulation_results}

We conduct experiments for DirectWAM and other baselines and methods mainly on two simulation benchmarks: LIBERO~\cite{libero} and RoboTwin 2.0~\cite{robotwin2}. We use 100 evaluation episodes per task for LIBERO. RoboTwin success rates are evaluated across all 50 tasks, using 50 episodes per task in Table~\ref{tab:main-experiment} and 10 episodes per task in all other success-rate evaluations. 

\newlength{\maintablewidth}
\setlength{\maintablewidth}{0.88\textwidth}

\paragraph{Learning strategy matters for single-step control performance.}
Table~\ref{tab:main-experiment} reveals a consistent progression from
naive truncation of cascade WAMs to learned future predictors.
On RoboTwin, simply reducing the iterative teacher to $1/1$ evaluations
causes success to fall to $54.9\%$.
Retaining additional computation in either stream partially recovers
performance, reaching $67.9\%$ for $1/2$ and $73.9\%$ for $2/1$, but both
remain below learned single-step models.
This shows that the remaining iterative computation cannot simply be
truncated; the direct mapping itself must be learned.

Among learned $1/1$ models, the training objective further separates
performance.
DirectWAM reaches $85.4\%$ average success on RoboTwin, compared with
$78.2\%$ for Flash-WAM.
The gap increases from $4.9$ percentage points on the Clean split
($87.0\%$ vs.\ $82.1\%$) to $9.5$ points under randomized evaluation
($83.7\%$ vs.\ $74.2\%$).
A similar numerical trend appears on LIBERO, although the Flash-WAM
results there are taken from the original paper rather than from a matched
checkpoint.
Both methods compress the cascade to the same single-step deployment
structure, but differ in how that mapping is learned:
Flash-WAM distills the pretrained generative process, whereas DirectWAM
evaluates the student-predicted future through the downstream action it
induces.
The resulting gap supports our central premise that, for direct WAM
prediction, a future should be learned according to its usefulness to the
action model rather than representation or generative fidelity alone.


\begin{table*}[t] \centering \caption{ \textbf{Main simulation results for cascade WAMs.} Success rates (\%) on RoboTwin~2.0 and LIBERO; higher is better. The full iterative LingBot teacher is italicized. Steps are video/action steps; Full denotes 20 video steps for LIBERO and 25 for RoboTwin, following the respective LingBot-VA evaluation configurations. RoboTwin results use 50 episodes per task across all 50 tasks. \textbf{Bold} denotes the better result between learned single-step methods (Flash-WAM and DirectWAM).
The $^{\dagger}$ denotes LIBERO results quoted from Flash-WAM because no corresponding checkpoint was released; these entries are therefore not treated as a strictly matched reproduction.
} \label{tab:main-experiment}  \setlength{\tabcolsep}{3.2pt} \renewcommand{\arraystretch}{1.16} \resizebox{\maintablewidth}{!}{ \begin{tabular}{@{}lcrrrrrrrr@{}} \toprule \multirow{2}{*}{Method} & \multirow{2}{*}{\makecell{Steps\\(Video / Action)}} & \multicolumn{3}{c}{\textbf{RoboTwin 2.0}} & \multicolumn{5}{c}{\textbf{LIBERO}} \\ \cmidrule(lr){3-5} \cmidrule(l){6-10} & & Clean & Rand. & Average. & Long & Goal & Object & Spatial & Average. \\ \midrule \textit{Lingbot Baseline}
    & \textit{Full / 50} & \textit{91.2} & \textit{90.2} & \textit{90.7}
    & \textit{98.2} & \textit{93.0} & \textit{97.2} & \textit{98.0} & \textit{96.6} \\
\midrule
\multirow{3}{*}{Naive Step Reduction}
    & 2 / 1 & 76.2 & 71.5 & 73.9 & 97.0 & 90.8 & 96.8 & 96.6 & 95.3 \\
    & 1 / 2 & 67.8 & 68.0 & 67.9 & 90.8 & 82.8 & 94.0 & 94.2 & 90.5 \\
    & 1 / 1 & 57.3 & 52.4 & 54.9 & 88.6 & 77.8 & 87.8 & 91.2 & 86.4 \\
 \specialrule{0.45pt}{3pt}{3pt} Flash-WAM & 1 / 1 & 82.1 & 74.2 & 78.2
    & 95.8\rlap{$^{\dagger}$} & 96.0\rlap{$^{\dagger}$} & 92.6\rlap{$^{\dagger}$}
    & 96.0\rlap{$^{\dagger}$} & 95.1\rlap{$^{\dagger}$} \\ \textbf{DirectWAM} & 1 / 1 & \textbf{87.0} & \textbf{83.7} & \textbf{85.4} & \textbf{97.2} & \textbf{96.6} & \textbf{96.2} & \textbf{98.5} & \textbf{97.1} \\ \bottomrule \end{tabular}} \end{table*}

\begin{figure}[t]
  \centering
  \begin{minipage}{\textwidth}

  \newsavebox{\wamParetoPanel}
  \sbox{\wamParetoPanel}{%
    \includegraphics[width=0.49\linewidth]{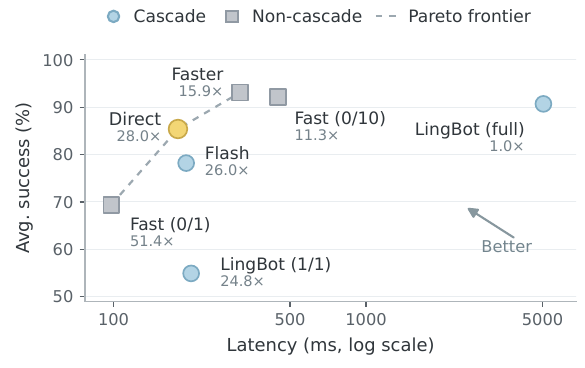}%
  }
  \captionsetup[subfigure]{
    labelformat=parens,
    labelsep=none,
    skip=2pt
  }

  \begin{subfigure}[t]{0.49\linewidth}
    \vspace{0pt}
    \centering
    \usebox{\wamParetoPanel}
    \caption{}
    \label{fig:robotwin_pareto}
  \end{subfigure}%
  \hfill
  \begin{subfigure}[t]{0.49\linewidth}
    \vspace{0pt}
    \centering
    \begin{minipage}[c][\ht\wamParetoPanel][c]{\linewidth}
      \centering
      \includegraphics[width=\linewidth]{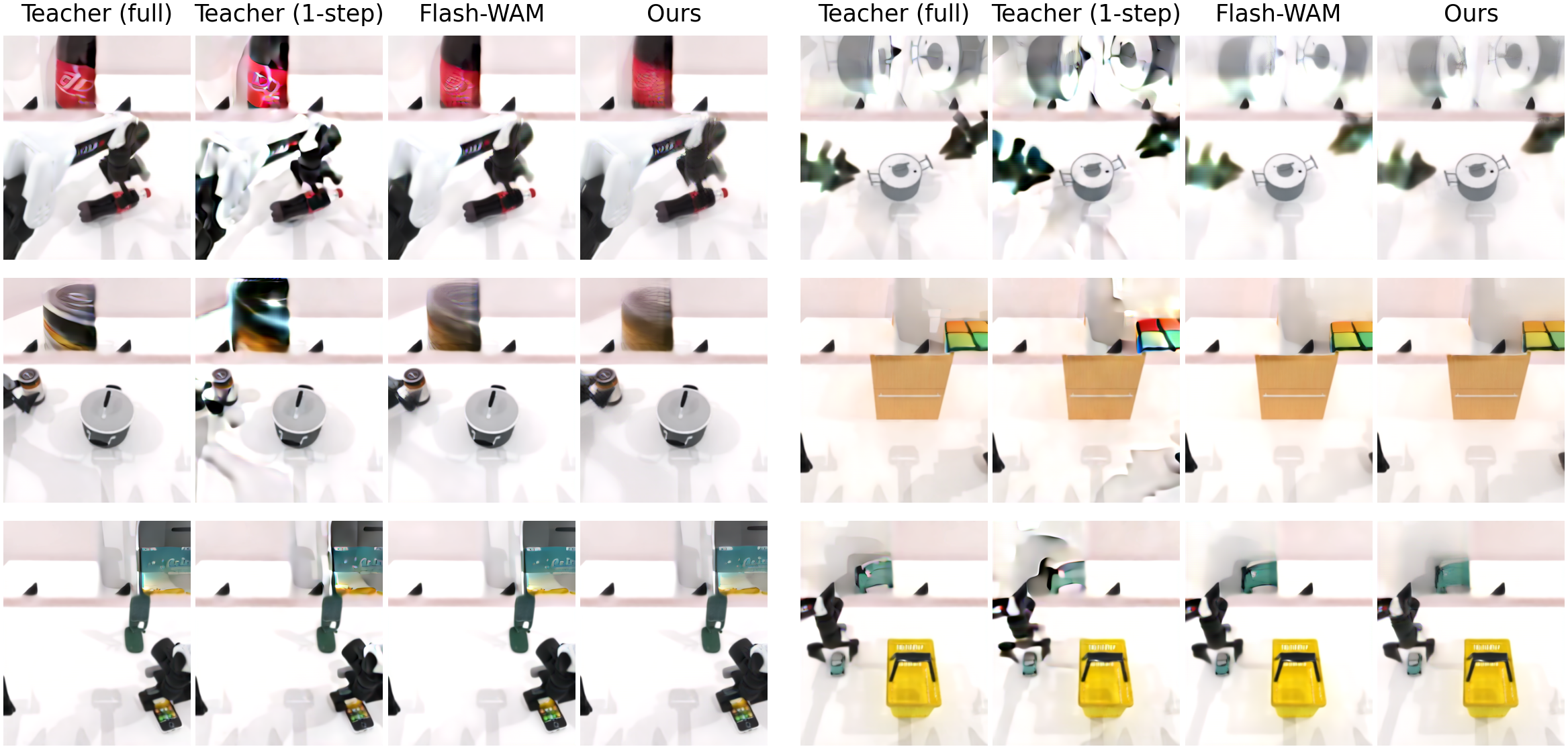}
    \end{minipage}
    \caption{}
    \label{fig:robotwin_selected_2x2}
  \end{subfigure}

  \end{minipage}

  \caption{
    \textbf{Efficiency and qualitative comparison on RoboTwin.}
    \textbf{(a)} Success--latency trade-off; labels show speedups over
    full LingBot-VA (5042 ms).
    \textbf{(b)} Six examples of predicted futures from the full teacher, the one-step
    teacher, Flash-WAM, and DirectWAM.
  }
  \label{fig:robotwin_efficiency_qualitative}
\end{figure}

\paragraph{DirectWAM remains on the efficiency--performance frontier.}
Figure~\ref{fig:robotwin_efficiency_qualitative} places DirectWAM in the
broader efficiency--performance landscape on RoboTwin.
Within cascade WAMs, DirectWAM substantially improves the trade-off under
compressed inference, achieving $85.4\%$ average success with a
$28.0\times$ reduction in model-side latency relative to the full iterative
teacher.

For broader context, we also include efficient WAMs with different
future-conditioning and deployment designs, such as FastWAM and FasterWAM\cite{fasterwam}.
We omit DIDO from this comparison because no released checkpoint is available
for a matched latency evaluation.
Despite these architectural differences, DirectWAM remains on the empirical
Pareto frontier among the evaluated configurations, occupying an intermediate
operating point that balances performance and efficiency.

More broadly, this result suggests that explicit future prediction and
iterative generative inference need not be coupled.
Retaining a future-to-action interface does not inherently require the
computational cost of multi-step diffusion.
DirectWAM preserves this interface while replacing iterative inference with
direct prediction, without moving off the observed efficiency--performance
frontier.

\paragraph{Action-supervised futures remain visually decodable.}
Using downstream action error as the primary supervision raises a natural
concern: could the intermediate future drift into an opaque code that is useful
only to the action model and sacrifice the explainability of cascade WAMs?
Figure~\ref{fig:robotwin_efficiency_qualitative} provides an empirical check on
this possibility by comparing decoded futures from different cascade WAM
variants on RoboTwin using predicted frames from different stages of three tasks.
Despite being trained primarily through action supervision, DirectWAM produces
recognizable future predictions that preserve the overall scene structure,
without the obvious visual collapse that would be expected from an unconstrained
latent code. Notably, DirectWAM exhibits fewer obvious ghosting artifacts and local color, texture, and shape distortions than the naively truncated one-step teacher.
This observation is consistent with the intended roles of the two objectives.
The action loss is free to reshape the future representation according to its
downstream control consequences, while the optional reconstruction term limits
drift from the demonstrated future.

Importantly, the stronger control performance of DirectWAM does not come with an
obvious visual degradation relative to Flash-WAM.
This suggests that improvement does not come simply from producing visually
better futures.
Rather, DirectWAM appears to improve how the predicted future supports the
downstream action model while preserving a decodable and visually meaningful
intermediate representation, retaining the explainability of cascade WAMs. We also provide a quantitative evaluation of predicted-future quality in Appendix~\ref{app:visual_prediction_quality}, showing that DirectWAM retains strong visual decodability despite being optimized for downstream control.

\subsection{Real-World Experiment}

\begin{figure}[!htbp]
    \centering
    \includegraphics[width=\linewidth]
        {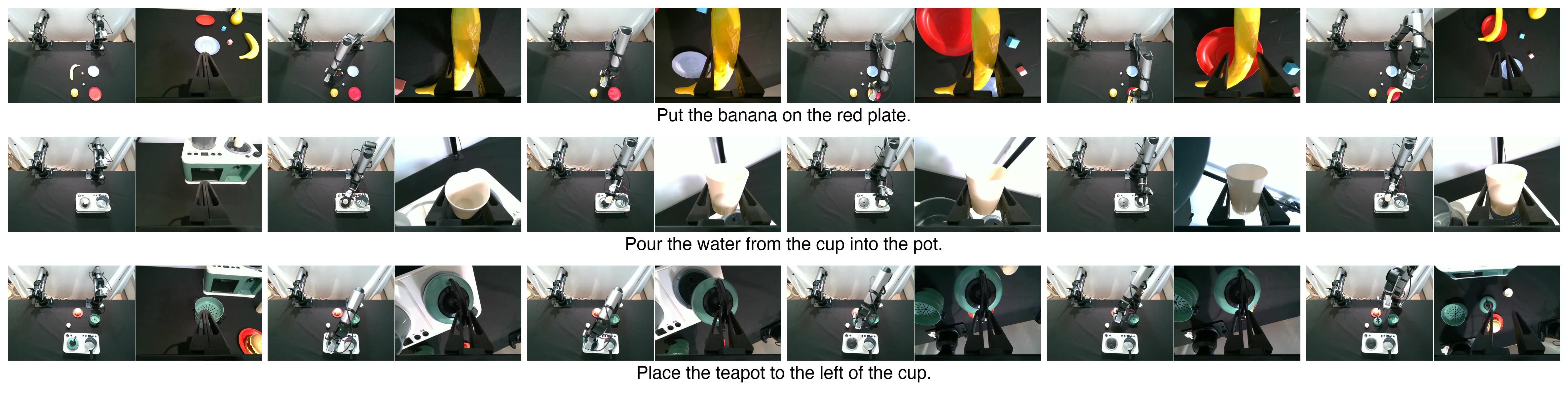}
    \caption{
        \textbf{Real-world task trajectories.}
        Six keyframes per task are ordered from left to right.
        Each keyframe pairs the external view (left)
        with the wrist view (right).
    }
    \label{fig:real_world_trajectories}
\end{figure}

\paragraph{Setting.}
Beyond simulation benchmarks, we evaluate whether DirectWAM preserves effective physical robot control. 
Experiments use a Galaxea A1 robotic arm equipped with a gripper, a fixed external camera (RealSense D435), and a wrist-mounted camera (RealSense D405).
We consider three tabletop manipulation tasks: Grasp Banana, Pouring Water, and Place Teapot, covering object transfer, pouring, and relative-position placement. We use 50 demonstration trajectories per task for training and compare DirectWAM with the LingBot-VA teacher and Flash-WAM over 20 evaluation trials per task. Figure~\ref{fig:real_world_trajectories} presents representative execution sequences from both camera views.

\paragraph{Results.}
DirectWAM preserves teacher-level performance in the real world while outperforming the single-step Flash-WAM baseline (Tab.~\ref{tab:real_world_sr}). All methods perform near saturation on Grasp Banana, whereas larger differences emerge on the more failure-prone tasks: DirectWAM achieves 11/20 successes on Pouring Water and 12/20 on Place Teapot, compared with 2/20 and 8/20 for Flash-WAM, respectively, while remaining comparable to the iterative teacher. Together with the simulation results, these findings support action-sufficient learning as an effective way to convert the iterative future-to-action cascade into direct prediction without sacrificing physical control.

\begin{table}[!htbp]
    \centering
    \caption{
        \textbf{Real-world task success rates.}
        Each method is evaluated over 20 trials per task.
        Results are reported as success rates (\%) with
        successful/total trials.
        Bold indicates the best result per task.
    }
    \label{tab:real_world_sr}

    \setlength{\tabcolsep}{6pt}
    \renewcommand{\arraystretch}{1.15}

    \begin{tabular}{@{}lccc@{}}
        \toprule
        Method & Grasp Banana & Pouring Water & Place Teapot \\
        \midrule
        LingBot-VA
        & \textbf{95 (19/20)}
        & 45 (9/20)
        & \textbf{60 (12/20)} \\

        DirectWAM
        & \textbf{95 (19/20)}
        & \textbf{55 (11/20)}
        & \textbf{60 (12/20)} \\

        Flash-WAM
        & 90 (18/20)
        & 10 (2/20)
        & 40 (8/20) \\
        \bottomrule
    \end{tabular}
\end{table}

\FloatBarrier
\section{Discussion}
\label{sec:discussion}

\paragraph{Separating pretrained capability from inference procedure.}
DirectWAM highlights a distinction between inheriting a diffusion-pretrained model and retaining its generative inference procedure. A pretrained future-to-action cascade can be converted into deterministic direct prediction without removing the intermediate future. This does not imply that iterative computation is redundant: naive step reduction causes substantial degradation. Rather, our results suggest that the useful capability acquired through diffusion pretraining can be re-expressed as direct computation better suited to downstream control. De-diffusion is therefore a model-conversion strategy that retains the predictive prior while replacing its deployment-time generative process.

\paragraph{Action sufficiency is relative to the downstream mapping.}
Our results suggest that the quality of a predicted future should be judged as part of the future-to-action interface, rather than by reconstruction fidelity alone. Prediction errors of similar representation-space magnitude can have very different action consequences, and training through the student-generated future directly exposes these consequences to the predictor. This does not imply that visual fidelity is unimportant or that there exists a fixed action-relevant subspace. Instead, for single-step cascade conversion, action sufficiency is defined relative to the direct action mapping that consumes the predicted future: the representation need not preserve every aspect of the reference future, but it must preserve what that downstream mapping requires for control.

\section{Conclusion}
\label{sec:conclusion}

We study whether a cascade World-Action Model can retain explicit future
prediction without retaining the stochastic, iterative diffusion process used
to realize it.
Our analysis shows that video-side sampling variation has limited downstream
effect, while directly truncating iterative inference leaves a substantial
control gap.
We further find that representation-space prediction error is an incomplete
proxy for control relevance, motivating action-sufficient future learning.
Based on these observations, we introduce \textbf{DirectWAM}, which converts
diffusion-pretrained video and action experts into deterministic single-step
predictors and trains the future representation through the action induced by
its own prediction.
DirectWAM preserves the future-to-action cascade while eliminating stochastic
sampling and iterative denoising at deployment, achieving strong control
performance with substantially reduced inference cost.
Overall, our results suggest that the predictive prior learned by diffusion
and the generative procedure used at inference need not be inseparable: for
robot control, future prediction can be re-learned as a direct,
action-sufficient computation.

\clearpage
\bibliographystyle{iclr2027_conference}
\bibliography{iclr2027_conference_verified}

\clearpage
\appendix

\section{Additional Analysis of Video-Side Stochasticity}
\label{app:video_stochasticity}

This appendix documents the analyses of stochastic future prediction.
We first examine how video-sampling variation affects a fixed action
model, then describe the clustering and closed-loop interventions.
Finally, we consolidate the additional multimodality diagnostic and its
controls.
These experiments address different questions: whether sampled futures
induce different actions, whether resampling benefits execution, and
whether detectable visual modes exist.
The control conclusions do not require the future distribution to be
unimodal.

\subsection{Paired Sampling and Offline Variation Propagation}
\label{app:sampling_metrics}

\paragraph{Sampling protocol.}
For each fixed demonstration context $h_i$, we generate $S$ future
predictions using independent video-noise seeds and evaluate them with
a common frozen single-step action model $\mathcal A$:
\begin{equation}
    z_i^{(s)} \sim p_{\theta}(\cdot \mid h_i),
    \qquad
    a_i^{(s)}
    =
    \mathcal A(z_i^{(s)},h_i;\bar\xi_{a,i}),
    \qquad s=1,\ldots,S.
    \label{eq:app_paired_sampling}
\end{equation}
The action-side input is held fixed across video samples within each
context.
The recorded RoboTwin analysis uses 237 contexts; LIBERO uses 64 contexts with 16 samples per context.
The model is the learned single-step head, not the teacher's native
multi-step action sampler.
This isolates the effect of replacing the future supplied to the same
model.

\paragraph{Distance and variance summaries.}
Let $D_V(i;s,s')$ and $D_A(i;s,s')$ denote the recorded distances
between two same-context future predictions and their decoded actions.
We report their Pearson correlation and mean pairwise distance ratio.
The latter depends on the representation and action scaling and should
not be interpreted as a scale-independent fraction of visual variation
transmitted to control.

To characterize sampling variation relative to contextual variation,
we also report, separately for video and action,
\begin{equation}
    R_X
    =
    \frac{W_X}{W_X+B_X},
    \qquad X\in\{V,A\},
    \label{eq:app_sampling_variance_fraction}
\end{equation}
where $W_X$ denotes within-context sampling variance and $B_X$
denotes between-context variance.
This is a fraction of total variance, not the ratio $W_X/B_X$.
The report additionally provides a median attenuation summary based
on per-context sampling RMS normalized by between-context RMS.
This RMS-based summary is distinct from the quotient $R_V/R_A$.

\begin{table}[htbp]
    \centering
    
    \setlength{\tabcolsep}{4pt}
    \caption{
    \textbf{Offline propagation of video-sampling variation.}
    All actions within a benchmark are evaluated with the same frozen
    single-step model.
    $R_V$ and $R_A$ are fractions of total variance.
    The final column reports the separately recorded median RMS
    attenuation.
    }
    \label{tab:app_sampling_propagation}
    \resizebox{\linewidth}{!}{%
    \begin{tabular}{@{}lcccccc@{}}
        \toprule
        Benchmark
        & Contexts $\times$ samples
        & $\rho(D_V,D_A)$
        & Mean $D_A/D_V$
        & $R_V$
        & $R_A$
        & RMS attenuation \\
        \midrule
        RoboTwin
        & $237\times32$
        & 0.69
        & 0.025
        & 6.6\%
        & 0.030\%
        & $20\times$ \\
        LIBERO
        & $64\times16$
        & 0.80
        & 0.138
        & 1.6\%
        & 0.25\%
        & $3\times$ \\
        \bottomrule
    \end{tabular}}
\end{table}

The positive correlations indicate that video-sampling variation is
reflected in the decoded actions, rather than being completely ignored.
Nevertheless, its relative contribution to action variance is small
compared with its contribution to video variance
(Table~\ref{tab:app_sampling_propagation}).
This is an offline observation under the chosen model; its
implications for execution are examined through the closed-loop
interventions below.

The propagation statistics above are reported for the original stores.
The multimodality diagnostic in Appendix~\ref{app:multimodality_test}
uses a separately audited 237-context RoboTwin subset.
These sample counts should not be treated as interchangeable.

\subsection{Clustering-Based Action-Separation Test}
\label{app:action_modes}

We complement the continuous summaries with a test of whether
representation-based groups induce different mean actions.
For each context, we apply $K$-means with $K=2$ and ten initializations
to the PCA-reduced future representations.
This imposes a two-way partition without assuming two natural modes.
For contexts flagged by the multimodality diagnostic, a two-component
Gaussian mixture model provides an additional partition.

For groups $\mathcal C_{i,1}$ and $\mathcal C_{i,2}$, we compute
\begin{equation}
    \bar a_{i,j}
    =
    \frac{1}{|\mathcal C_{i,j}|}
    \sum_{s\in\mathcal C_{i,j}}a_i^{(s)},
    \qquad
    S_{\mathrm{act},i}
    =
    \frac{
        \|\bar a_{i,1}-\bar a_{i,2}\|_2
    }{
        \|a_i^\star\|_2
    },
    \label{eq:app_cluster_separation}
\end{equation}
where $a_i^\star$ is the demonstrated action chunk.
The reported separation for the forced partitions is 0.2\% of the
demonstrated action magnitude.
This measures differences between cluster means, not equality of the
complete action distributions.
Its role is therefore supplementary: under this representation-based
partition, the two groups induce little mean-action separation.

\subsection{Closed-Loop Sampling Interventions}
\label{app:closed_loop_sampling}

\paragraph{Protocol and interventions.}
We evaluate both the iterative teacher and the single-step checkpoint
on RoboTwin Clean using 50 tasks and ten episodes per task
($n=500$ per setting).
Scene initialization and instruction settings are shared across
variants, and the action-noise schedule is paired.
Unlike the common-model offline analysis, these evaluations retain
each policy's own action pathway.

Standard sampling draws fresh video noise at each control chunk.
\emph{Fixed noise} reuses one in-distribution realization throughout
an episode, evaluated separately for seeds 123, 456, and 789.
\emph{Avg-4} averages four independently predicted future latents
at each chunk before action inference:
\begin{equation}
    \xi_{v,t}=\bar\xi_v
    \quad\text{(Fixed noise)},
    \qquad
    \bar z_t=\frac14\sum_{m=1}^{4}z_t^{(m)}
    \quad\text{(Avg-4)}.
    \label{eq:app_sampling_interventions}
\end{equation}
These are input interventions on the same checkpoint, not separately
trained models.
Fixing a sampled noise tensor is distinct from replacing it with a
zero tensor.

\begin{table}[htbp]
    \centering
    
    \setlength{\tabcolsep}{5pt}
    \caption{
    \textbf{Video-sampling interventions on RoboTwin Clean.}
    Success rates are percentages over 500 episodes per entry (50 tasks, 10 episodes per task).
    Fixed-noise mean is the arithmetic mean of the three seed-specific
    success rates.
    }
    \label{tab:app_sampling_interventions}
    \resizebox{\linewidth}{!}{%
    \begin{tabular}{@{}lrrrrrr@{}}
        \toprule
        Model
        & Standard
        & Fixed 123
        & Fixed 456
        & Fixed 789
        & Fixed-noise mean
        & Avg-4 \\
        \midrule
        Iterative teacher
        & 91.6 & 92.4 & 91.6 & 89.6 & 91.2 & 92.0 \\
        Single-step model
        & 87.0 & 87.4 & 87.6 & 89.4 & 88.1 & 87.0 \\
        \bottomrule
    \end{tabular}}
\end{table}

The reported paired tests give absolute success-rate changes of at
most 2.4 percentage points across the eight interventions, with
McNemar $p\geq0.12$ and bootstrap 95\% confidence intervals
containing zero.
For example, seed 789 changes teacher success by $-2.0$ points
($[-4.4,0.4]$) but student success by $+2.4$ points
($[-0.4,5.2]$).
These results show no detected systematic benefit of resampling
under the evaluated protocol; they do not establish exact
performance equivalence.
Individual episode outcomes can change even when aggregate success
remains similar.

The recorded standard-sampling baseline uses an unseeded global video
RNG, so pairing concerns episodes and action noise rather than
identical video-noise realizations.
The teacher Avg-4 checkpoint was matched by recorded provenance
rather than a stored hash.

\paragraph{Repeated trajectories.}
We additionally evaluate 30 scene--instruction contexts from 15 tasks,
using two scene seeds per task selected for stronger single-chunk
multimodality evidence.
Each context is executed eight times from the same initial state,
with paired action-noise schedules and varying video noise.
This targeted set is not a random benchmark sample.

Let $x_s(t)\in\mathbb R^3$ be the commanded position of the moving
end effector in rollout $s$.
Trajectories are truncated to their common prefix,
$T=\min_s T_s$.
We compute the mean path and time-averaged RMS spread as
\begin{equation}
    \bar x(t)=\frac1S\sum_{s=1}^{S}x_s(t),
    \qquad
    d=\frac1T\sum_{t=1}^{T}
       \sqrt{
           \frac1S\sum_{s=1}^{S}
           \|x_s(t)-\bar x(t)\|_2^2
       },
    \quad S=8.
    \label{eq:app_trajectory_spread}
\end{equation}
The normalized spread is
\begin{equation}
    R_s
    =
    \left\|
        \max_{1\leq t\leq T}x_s(t)
        -
        \min_{1\leq t\leq T}x_s(t)
    \right\|_2,
    \qquad
    R=\operatorname{median}_s R_s,
    \qquad
    D_{\mathrm{traj}}=d/R,
    \label{eq:app_trajectory_normalization}
\end{equation}
where maxima and minima are coordinatewise.

In 25 contexts, all eight rollouts succeed; four contexts have mixed
outcomes, and one fails in all repetitions.
The median normalized spread is $2\%$ across all 30 contexts and remains $2\%$ within the 25 all-success contexts. For these 25 contexts, the median absolute spread is $0.86\,\mathrm{cm}$ and the median motion range is $43\,\mathrm{cm}$.
These are separately aggregated statistics, not a conversion obtained
by dividing the two medians.
The metric summarizes commanded positions over the shared prefix;
it does not measure full action-distribution equivalence or all
aspects of execution.

\subsection{Multimodality Analysis of Sampled Futures}
\label{app:multimodality_test}

This diagnostic asks whether repeated future samples exhibit
detectable multimodal structure.
It supplements, rather than establishes a premise for, the action
and closed-loop analyses above.

\paragraph{Data and held-out representation.}
We use the RoboTwin teacher with 25 video denoising steps and
guidance scale 5, generating 32 samples per context.
The original store contains 256 contexts from 50 tasks.
Nineteen contexts have mismatched history latents and prompts due
to an indexing issue; the multimodality analysis therefore uses the
remaining 237 contexts from 47 tasks.
Each future has shape $48\times2\times24\times20$.
Averaging the two latent frames and applying $4\times4$ spatial
average pooling produces a 1,440-dimensional vector per sample.

We split contexts into two disjoint sets, fit a PCA basis to
sampling deviations in one set, and project samples in the other;
the roles are then exchanged.
For context $i$,
\begin{equation}
    y_i^{(s)}=\phi_{-i}(z_i^{(s)}),
    \label{eq:app_crossfit_pca}
\end{equation}
where $\phi_{-i}$ is fitted without context $i$.
We retain $d=8$ principal components, with $d=16$ as a robustness
check.
This prevents a tested context from determining its own PCA basis.

\paragraph{Projection-based testing.}
We evaluate 28 predefined one-dimensional projections per context.
In the $d=8$ setting, these comprise the eight PCA coordinate axes
and twenty fixed random unit directions.
The directions are shared across real samples and the corresponding
controls.
For direction $u_m$, define
\begin{equation}
    q_{i,s}^{(m)}=u_m^\top y_i^{(s)},
    \qquad
    H_{i,m}(x)
    =
    \frac1S\sum_{s=1}^{S}
    \mathbf 1[q_{i,s}^{(m)}\leq x].
\end{equation}
Hartigan's dip statistic~\cite{hartigan_dip} is
\begin{equation}
    D_{i,m}
    =
    \inf_{G\in\mathcal F_{\mathrm{uni}}}
    \|H_{i,m}-G\|_{\infty},
    \label{eq:app_dip_statistic}
\end{equation}
where $\mathcal F_{\mathrm{uni}}$ is the family of unimodal CDFs.
Let $p_{i,m}$ be the $p$-value returned by the dip test under its
unimodal null.
A context is flagged when
\begin{equation}
    \min_{m=1,\ldots,28}p_{i,m}<0.05/28,
    \label{eq:app_dip_threshold}
\end{equation}
using a Bonferroni correction across projections.

\paragraph{Calibration and positive controls.}
For the matched-unimodal check, we estimate the mean $\mu_i$ and
covariance $\Sigma_i$ in PCA space and draw matched Gaussian samples,
\begin{equation}
    \tilde y_i^{(s)}
    \sim
    \mathcal N(\mu_i,\Sigma_i),
    \qquad s=1,\ldots,32.
\end{equation}
The existing calibration report contains 1,280 null sets, generated
as five replicates per reference context.
The reported detection rates are 0.16\% for PCA-8 and 0\% for PCA-16.

Positive controls combine 16 samples from each of two different
contexts of the same task.
We evaluate 190 such two-context mixtures using the same held-out
representation and corrected testing procedure.
Detection rates are 97.4\% for PCA-8 and 99.5\% for PCA-16.
These controls demonstrate sensitivity to the constructed balanced
mixtures, not to every possible form of multimodality.

\begin{table}[htbp]
    \centering
    
    \setlength{\tabcolsep}{6pt}
    \caption{
    \textbf{Multimodality diagnostic and controls.}
    Real-data entries give detected contexts over the 237-context
    audited set.
    Control entries report detection rates.
    }
    \label{tab:future_multimodality}
    \begin{tabular}{@{}lcc@{}}
        \toprule
        Sample set & PCA-8 & PCA-16 \\
        \midrule
        Same-context futures
        & $0/237$ (0.0\%) & $1/237$ (0.4\%) \\
        Matched Gaussian controls
        & 0.16\% & 0.0\% \\
        Two-context mixtures
        & 97.4\% & 99.5\% \\
        \bottomrule
    \end{tabular}
\end{table}

\paragraph{Interpretation and limitations.}
We find little evidence of multimodality at the tested representation
resolution and sample count.
The sole PCA-16 detection is not consistently separated across the
reported checks.
Non-detection does not establish global unimodality: discarded
representation directions may contain modes, 32 samples provide
limited coverage of rare modes, and balanced positive controls are
easier than highly imbalanced mixtures.
Moreover, this diagnostic uses one predicted chunk under
teacher-forced history.
Its conclusions need not describe the full state distribution
reached in closed-loop execution.

\clearpage
\section{Training Details}
\label{app:training_details}

Both benchmarks initialize from the released LingBot-VA post-trained
checkpoints.
The documented two-stage training runs use fully sharded data parallelism,
a per-GPU batch size of one, and
AdamW~\cite{adamw} with $\beta=(0.9,0.95)$ and weight decay 0.1.

\paragraph{Stage 1: single-step action learning.}
The video expert is frozen while the action expert is trained against
demonstrated action chunks.
Supervision is applied to the final action prediction.
On LIBERO-Long, this stage uses 600 optimization steps and an
effective batch size of 16.
On RoboTwin, it uses 8,000 steps at learning rate $10^{-5}$,
taking approximately 3.5 hours.

\paragraph{Stage 2: action-sufficient future learning.}
The future predictor is trained through the action decoded from its
own predicted future, with the action expert frozen during the
plotted training phase.
LIBERO-Long uses 400 steps, an effective batch size of eight,
and no reconstruction anchor ($\lambda_v=0$); the two stages
together take approximately 21 minutes.
RoboTwin uses 6,500 steps at learning rate $10^{-5}$ with
$\lambda_v=0.01$, taking approximately six hours for Stage 2.
The training curves and their interpretation are given in
Appendix~\ref{app:stage2_dynamics}.

\newpage
\section{Additional Analysis of Action-Sufficient Learning}
\label{app:action_sufficient_analysis}

We examine the learned future predictor from two complementary
perspectives.
Training curves describe how action and reconstruction objectives
evolve during Stage 2.
Selective replacement then tests which components of the difference
between the learned predictors account for their downstream action
improvement, first through offline action error and then through
closed-loop control.

\subsection{Stage-2 Training Dynamics}
\label{app:stage2_dynamics}

\begin{figure}[htbp]
    \centering
    \includegraphics[width=0.90\linewidth]
    {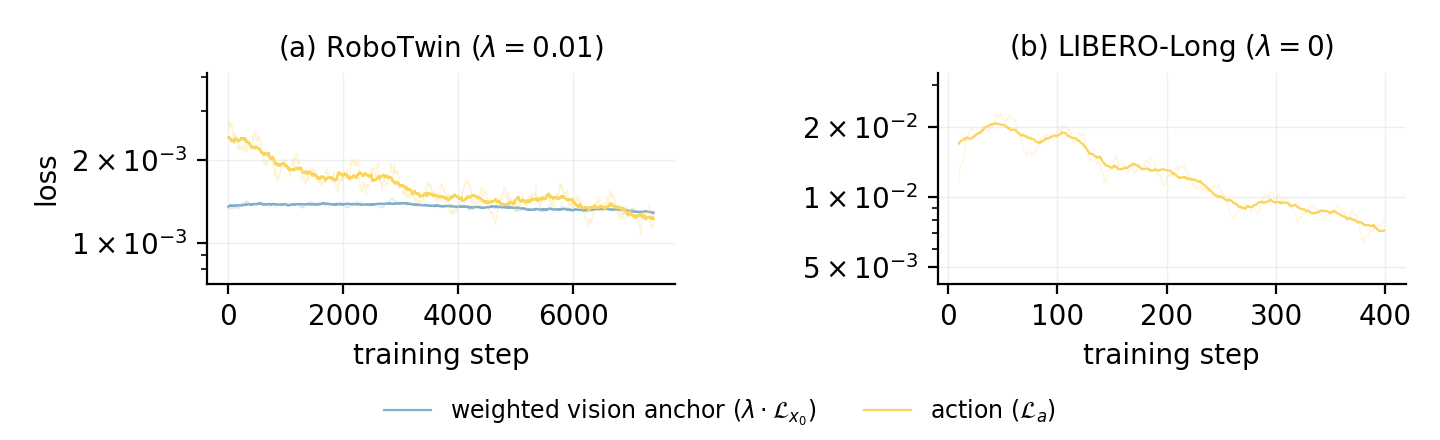}
    \caption{
    \textbf{Stage-2 training dynamics.}
    The action expert is frozen during the plotted phase.
    RoboTwin (left) uses $\lambda_v=0.01$ and shows both action loss
    and the weighted future-anchor loss.
    LIBERO-Long (right) uses $\lambda_v=0$ and shows only action loss.
    The vertical scale is logarithmic.
    }
    \label{fig:training_curves}
\end{figure}

Figure~\ref{fig:training_curves} reports the phase in which the
action expert is held fixed, and the video expert is updated.
We first consider LIBERO-Long, where $\lambda_v=0$.
Despite receiving no reconstruction supervision, the future-prediction
pathway learns to reduce the action loss over training.
The reduction therefore does not require updating the action expert
or explicitly minimizing a video reconstruction objective.

RoboTwin provides a complementary setting with $\lambda_v=0.01$.
Here, action loss decreases substantially while the weighted anchor
loss remains approximately stable.
Improved action regression is therefore not accompanied by a
comparable reduction in the measured reconstruction error.
This is consistent with using reconstruction as an anchor while
evaluating future usefulness through the action objective.

These curves characterize training objectives, not success rates.
They do not establish that visual quality is unchanged, that the
anchor has no optimization effect, or that all action improvement
lies in a particular subspace.
The next experiment examines the last question directly in the
learned predictions.

\subsection{WHICH LEARNED PREDICTION DIFFERENCES MATTER FOR CONTROL?}
\label{app:student_sensitive_replacement}

\paragraph{Comparison and intervention.}
We evaluate 64 RoboTwin contexts with demonstrated history.
Let $z_i^{\mathrm V}$ be the vision-supervised student's future
and $z_i^{\mathrm A}$ DirectWAM's future.
Both are decoded by the same frozen single-step action model
$\mathcal A$ with paired action-side inputs.
This comparison holds the model fixed while replacing the
predicted future.

For each context, let $J_i$ be the local Jacobian of this model
at $z_i^{\mathrm V}$.
We map 32 sampled action-output directions back to future space
and orthonormalize them:
\begin{equation}
    Q_i
    =
    \operatorname{orth}
    \left(
        [J_i^\top u_1,\ldots,J_i^\top u_{32}]
    \right),
    \qquad
    P_i=Q_iQ_i^\top.
    \label{eq:app_student_probe}
\end{equation}
This is a context-local probe of action-sensitive directions,
not a claim to recover the complete sensitive subspace or the
leading singular vectors.
We selectively transfer the learned prediction difference,
\begin{equation}
    \Delta z_i
    =
    z_i^{\mathrm A}-z_i^{\mathrm V},
    \qquad
    z_i^{\mathrm{sens}}
    =
    z_i^{\mathrm V}+P_i\Delta z_i.
    \label{eq:app_student_repair}
\end{equation}
Unlike the teacher-referenced correction in the main text,
this intervention moves between two learned single-step predictions.
Its full-replacement reference is $z^{\mathrm A}$, not the teacher.

\paragraph{Evaluation.}
Let $\mathcal E(q)$ be the aggregate offline distance between the
actions decoded from candidate futures $q$ and the demonstrated
actions, using the evaluated action coordinates.
The report uses an $L_2$ action distance, not a closed-loop
success metric.
We define pooled recovery as
\begin{equation}
    \operatorname{Recovery}(q)
    =
    \frac{
        \mathcal E(z^{\mathrm V})-\mathcal E(q)
    }{
        \mathcal E(z^{\mathrm V})-\mathcal E(z^{\mathrm A})
    }.
    \label{eq:app_student_recovery}
\end{equation}
Recovery is computed from aggregate errors before forming the ratio.
It is distinct from a median of per-context recovery ratios.
Equal-norm random-subspace and Gaussian perturbations provide
controls for correction magnitude.

\begin{table}[htbp]
    \centering
    
    \setlength{\tabcolsep}{6pt}
    \caption{
    \textbf{Selective replacement of learned prediction differences.}
    Results use 64 RoboTwin contexts and the same frozen action model.
    The selected component accounts for 1.9\% of the
    prediction-difference energy and recovers approximately 80\%
    of the pooled offline action-error improvement.
    Dashes indicate that only the recovery summary is reported
    for the random controls.
    }
    \label{tab:student_sensitive_replacement}
    \begin{tabular}{@{}lcc@{}}
        \toprule
        Prediction
        & GT-action error $\downarrow$
        & Recovery $\uparrow$ \\
        \midrule
        Vision-supervised $z^{\mathrm V}$
        & 0.257 & 0\% \\
        Full replacement $z^{\mathrm A}$
        & 0.216 & 100\% \\
        Selective replacement $z^{\mathrm{sens}}$
        & 0.224 & 80\% \\
        Equal-norm random controls
        & -- & 1--2\% \\
        \bottomrule
    \end{tabular}
\end{table}

\begin{table}[t]
\centering

\setlength{\tabcolsep}{5pt}
\caption{
\textbf{Closed-loop selective replacement of learned prediction differences.}
We keep the same downstream action model fixed and vary only the
future representation supplied to it.
The selective correction transfers only the locally action-sensitive
component of the difference between the vision-supervised and
DirectWAM futures.
}
\label{tab:closed_loop_selective_replacement}
\begin{tabular}{@{}lcc@{}}
\toprule
Future representation
& Transferred energy
& Success rate (\%) $\uparrow$ \\
\midrule
Vision-supervised $z^V$
& 0\% & 78.2 \\
Selective replacement $z^V + P(z^A-z^V)$
& 1.4\% & \textbf{87.3} \\
DirectWAM $z^A$
& 100\% & 87.0 \\
\bottomrule
\end{tabular}
\end{table}

\paragraph{Results and scope.}
Table~\ref{tab:student_sensitive_replacement} shows that the effect
of the learned prediction difference is highly non-uniform under
the frozen action model.
Full replacement reduces the offline demonstration-action error
from 0.257 to 0.216, while selectively transferring a locally
action-sensitive component containing only 1.9\% of the
prediction-difference energy reduces the error to 0.224,
recovering approximately 80\% of the full offline improvement.
Equal-norm random controls recover only 1--2\%, indicating that
the benefit depends strongly on the direction of the correction
rather than its magnitude alone.
We further test whether this effect persists in closed-loop control.
As shown in Table~\ref{tab:closed_loop_selective_replacement},
starting from the vision-supervised future, transferring only
the locally action-sensitive component of the learned DirectWAM
difference, corresponding to 1.4\% of the difference energy,
increases RoboTwin success from 78.2\% to 87.3\%, essentially
matching the 87.0\% obtained by full replacement with the
DirectWAM future.
Together, these interventions show that the downstream consequences
of the learned prediction difference are highly non-uniform:
a small locally action-sensitive component is sufficient to recover
a disproportionate share of the offline action improvement and
essentially the full closed-loop control gain.
These results do not imply that the complete learned correction
is concentrated in a fixed action-relevant subspace; rather,
they show that a small local component can affect its
downstream control consequences.

\subsection{Teacher-Referenced Action-Sensitive Error Recovery}
\label{app:teacher_sensitive_recovery}
This section details the diagnostic in Fig.~\ref{fig:action_sensitive_recovery}.
It compares selective corrections toward a full-teacher future for the vision-supervised student, Flash-WAM, and naive one- and two-step predictors.
The reference is the teacher prediction, rather than the demonstrated future or the output of another learned student.

\paragraph{Local projection.}
Let $\mathcal A(z,h)$ denote the frozen action model for conditioning context $h$ and future representation $z$.
At the base prediction $z^{\mathrm B}$, we form the local Jacobian $J=\left.\partial\mathcal A(z,h)/\partial z\right|_{z=z^{\mathrm B}}$.
We map $r=32$ sampled action-output directions $u_i$ back into representation space and orthonormalize the resulting vectors:
\begin{equation}
Q=\operatorname{orth}\!\left([J^\top u_1,\ldots,J^\top u_r]\right),
\qquad P=QQ^\top.
\label{eq:action_sensitive_projection}
\end{equation}
This Jacobian-based construction supplies a context-local diagnostic subspace of rank at most 32.
It does not identify the complete action-sensitive subspace or necessarily recover the leading singular vectors of $J$.

\paragraph{Selective correction.}
For $\Delta z=z^{\mathrm T}-z^{\mathrm B}$, the two components in Eq.~\ref{eq:error_decomposition} define candidate futures
\begin{equation}
z^{\mathrm{sens}}=z^{\mathrm B}+P\Delta z,
\qquad
z^{\mathrm{comp}}=z^{\mathrm B}+(I-P)\Delta z.
\label{eq:app_teacher_candidates}
\end{equation}
Each candidate is decoded separately using the frozen model.
The full correction yields $z^{\mathrm B}+\Delta z=z^{\mathrm T}$ and provides the reference improvement.
Because $P$ is an orthogonal projector, the squared discrepancy decomposes as $\|\Delta z\|_2^2=\|P\Delta z\|_2^2+\|(I-P)\Delta z\|_2^2$.
The sensitive energy fraction is therefore $\|P\Delta z\|_2^2/\|\Delta z\|_2^2$ for a nonzero discrepancy.
Here, discrepancy refers to distance from the teacher prediction, not necessarily error relative to the true future.

\paragraph{Action-error recovery.}
Let $\mathcal E(q)$ denote the offline error of the actions decoded from candidate futures $q$ relative to demonstrated actions, with the same evaluation cases and error aggregation used for the base and teacher references.
The normalized recovery compares the candidate's improvement with that obtained by full teacher replacement:
\begin{equation}
\operatorname{Recovery}(q)=
\frac{\mathcal E(z^{\mathrm B})-\mathcal E(q)}
{\mathcal E(z^{\mathrm B})-\mathcal E(z^{\mathrm T})}.
\label{eq:app_teacher_recovery}
\end{equation}
The denominator is the available teacher-referenced improvement, not the total base action error; interpreting the ratio as recovery presumes a positive reference improvement.
A value of $100\%$ means matching the teacher-reference action error, rather than eliminating all action error.
The metric measures offline action prediction, not closed-loop success.
Recoveries from the two separate corrections need not sum to $100\%$, because the model and the action-error functional can respond nonlinearly to finite changes in the future representation.

\paragraph{Interpretation and scope.}
Sensitivity is favored by the construction of $P$.
What is measured is whether correcting the actual teacher--prediction discrepancy along those directions reduces demonstration-action error, and by how much; local sensitivity alone does not specify that finite-correction outcome.
Across the four predictors, the selected component contains $0.31$--$0.85\%$ of discrepancy energy and recovers $16.3$--$34.0\%$ of the available action-error improvement.
For the vision-supervised student, the corresponding comparison is $0.85\%$ energy and $34.0\%$ recovery for the selected component, versus $99.15\%$ energy and $24.8\%$ recovery for its complement.
The complement remains useful and is not an action-insensitive nullspace.
These observations support using downstream action consequences in addition to representation distance; they do not establish that this probe explains the entire closed-loop performance gap or that learning should be restricted to its subspace.
The projection is used only for diagnosis and is not required by the training objective.

For scale, an independent uniformly random rank-$r$ subspace in a $d$-dimensional representation captures an expected fraction $r/d$ of the energy of a fixed discrepancy, approximately $0.07\%$ for rank 32 in this representation.
This is a geometric reference for energy concentration, not an empirical random-correction baseline for action-error recovery.

\clearpage
\subsection{Visual Prediction Quality}
\label{app:visual_prediction_quality}
Setting. We evaluate predicted future videos offline under a teacher-forced protocol, where ground-truth history is committed to the KV cache and each model predicts the next future chunk independently. On RoboTwin, we evaluate 4,739 valid contexts sampled from the 2,500 clean demonstrations. We compare multi-step teacher variants, the one-step teacher, Flash-WAM, and DirectWAM, using the same random seed and benchmark-specific guidance settings. Predicted and ground-truth latent chunks are decoded independently with the Wan2.2 VAE, and metrics are computed per camera and then averaged. We report PSNR, SSIM, LPIPS, latent MSE, and distribution-level FID/KID; confidence intervals are computed over contexts.

\begin{table}[htbp]
\centering

\setlength{\tabcolsep}{9pt}
\caption{
\textbf{Visual prediction quality.}
Teacher variants use guidance scale 5 with the indicated number
of video denoising steps. Flash-WAM and DirectWAM use one step.
Bold indicates the best value among the listed methods.
}
\label{tab:visual_prediction_quality}
\begin{tabular}{@{}lccccc@{}}
\toprule
Method
& PSNR $\uparrow$
& SSIM $\uparrow$
& LPIPS $\downarrow$
& FID $\downarrow$
& KID $\times 10^3$ $\downarrow$ \\
\midrule
Teacher, 25 steps & 30.37 & 0.938 & \textbf{0.051} & \textbf{2.8} & \textbf{0.56} \\
Teacher, 8 steps  & 30.16 & 0.936 & 0.054 & 3.5 & 0.93 \\
Teacher, 4 steps  & 29.64 & 0.932 & 0.060 & 5.0 & 1.78 \\
Teacher, 2 steps  & 28.57 & 0.920 & 0.075 & 8.1 & 3.74 \\
Teacher, 1 step   & 25.80 & 0.885 & 0.123 & 17.4 & 10.24 \\
\midrule
Flash-WAM        & 26.62 & 0.928 & 0.077 & 10.5 & 4.11 \\
DirectWAM (ours) & \textbf{31.79} & \textbf{0.949} & 0.052 & 5.9 & 2.06 \\
\bottomrule
\end{tabular}
\end{table}

DirectWAM retains high-quality visual predictions despite being trained primarily for downstream control. As shown in Tab.~\ref{tab:visual_prediction_quality}, directly truncating the diffusion teacher progressively degrades future prediction quality, with the one-step teacher dropping to 25.80 PSNR, 0.885 SSIM, and 0.123 LPIPS. Flash-WAM partially recovers this degradation, while DirectWAM achieves the strongest reference-based reconstruction quality among the single-step methods, reaching 31.79 PSNR, 0.949 SSIM, and 0.052 LPIPS. Notably, its LPIPS is nearly identical to the full 25-step teacher (0.051), although distribution-level metrics remain worse than the full teacher (FID 5.9 vs. 2.8; KID 2.06 vs. 0.56). These results indicate that optimizing the predicted future through downstream action supervision does not collapse the intermediate representation into an opaque action code: DirectWAM preserves substantial visual structure while converting the iterative future predictor into a single-step mapping.

\end{document}

%% file: math_commands.tex
\usepackage{amsmath,amsfonts,bm}

\def\eqref#1{equation~\ref{#1}}

\def\1{\bm{1}}

\DeclareMathAlphabet{\mathsfit}{\encodingdefault}{\sfdefault}{m}{sl}
\SetMathAlphabet{\mathsfit}{bold}{\encodingdefault}{\sfdefault}{bx}{n}



%% file: iclr2027_conference_verified.bib
@article{lingbotva,
  title={Causal World Modeling for Robot Control},
  author={Li, Lin and Zhang, Qihang and Luo, Yiming and Yang, Shuai and Wang, Ruilin and Han, Fei and Yu, Mingrui and Gao, Zelin and Xue, Nan and Zhu, Xing and Shen, Yujun and Xu, Yinghao},
  journal={arXiv preprint arXiv:2601.21998},
  year={2026}
}

@inproceedings{unipi,
 author = {Du, Yilun and Yang, Sherry and Dai, Bo and Dai, Hanjun and Nachum, Ofir and Tenenbaum, Josh and Schuurmans, Dale and Abbeel, Pieter},
 booktitle = {Advances in Neural Information Processing Systems},
 doi = {10.52202/075280-0403},
 editor = {A. Oh and T. Naumann and A. Globerson and K. Saenko and M. Hardt and S. Levine},
 pages = {9156--9172},
 publisher = {Curran Associates, Inc.},
 title = {Learning Universal Policies via Text-Guided Video Generation},
 url = {https://proceedings.neurips.cc/paper_files/paper/2023/file/1d5b9233ad716a43be5c0d3023cb82d0-Paper-Conference.pdf},
 volume = {36},
 year = {2023}
}

@misc{dreamzero,
      title={World Action Models are Zero-shot Policies},
      author={Seonghyeon Ye and Yunhao Ge and Kaiyuan Zheng and Shenyuan Gao and Sihyun Yu and George Kurian and Suneel Indupuru and You Liang Tan and Chuning Zhu and Jiannan Xiang and Ayaan Malik and Kyungmin Lee and William Liang and Nadun Ranawaka and Jiasheng Gu and Yinzhen Xu and Guanzhi Wang and Fengyuan Hu and Avnish Narayan and Johan Bjorck and Jing Wang and Gwanghyun Kim and Dantong Niu and Ruijie Zheng and Yuqi Xie and Jimmy Wu and Qi Wang and Ryan Julian and Danfei Xu and Yilun Du and Yevgen Chebotar and Scott Reed and Jan Kautz and Yuke Zhu and Linxi "Jim" Fan and Joel Jang},
      year={2026},
      eprint={2602.15922},
      archivePrefix={arXiv},
      primaryClass={cs.RO},
      url={https://arxiv.org/abs/2602.15922},
}

@InProceedings{motus,
    author    = {Bi, Hongzhe and Tan, Hengkai and Xie, Shenghao and Wang, Zeyuan and Huang, Shuhe and Liu, Haitian and Zhao, Ruowen and Feng, Yao and Xiang, Chendong and Rong, Yinze and Zhao, Hongyan and Liu, Hanyu and Su, Zhizhong and Ma, Lei and Su, Hang and Zhu, Jun},
    title     = {Motus: A Unified Latent Action World Model},
    booktitle = {Proceedings of the IEEE/CVF Conference on Computer Vision and Pattern Recognition (CVPR)},
    month     = {June},
    year      = {2026},
    pages     = {35101-35113}
}

@article{fastwam,
  title={Fast-WAM: Do World Action Models Need Test-time Future Imagination?},
  author={Tianyuan Yuan and Zibin Dong and Yicheng Liu and Hang Zhao},
  journal={arXiv preprint arXiv:2603.16666},
  year={2026},
  url={https://arxiv.org/abs/2603.16666}
}

@article{fasterwam,
  title   = {Faster-WAM: Efficient Inference-Time Future Conditioning for Robust World Action Models},
  author  = {Zhao, Weiheng and Jiang, Haoyi and Shi, Xin and Liu, Liu and Huang, Fan and Su, Zhizhong and Sui, Wei and Wang, Xinggang},
  journal = {arXiv preprint arXiv:2608.04404},
  year    = {2026}
}

@misc{flashwam,
      title={Flash-WAM: Modality-Aware Distillation for World Action Models}, 
      author={Arman Akbari and Ci Zhang and Arash Akbari and Lin Zhao and Yixiao Chen and Weiwei Chen and Xuan Zhang and Geng Yuan and Yanzhi Wang},
      year={2026},
      eprint={2606.05254},
      archivePrefix={arXiv},
      primaryClass={cs.LG},
      url={https://arxiv.org/abs/2606.05254}, 
}

@misc{dido,
      title={DIDO: Distilling Interaction-Centric Dynamics into One-Step Denoising for World Action Models}, 
      author={Jing Lyu and Shuanghao Bai and Runze Xiao and Zhenyu Liao and Wenxing Tan and Zihan Tang and Ruochuan Shi and Cheng Peng and Yuheng Ji and Yihao Wang and Badong Chen and Pengwei Wang and Zhongyuan Wang and Xiaoguang Zhao},
      year={2026},
      eprint={2609.15570},
      archivePrefix={arXiv},
      primaryClass={cs.RO},
      url={https://arxiv.org/abs/2609.15570}, 
}

@InProceedings{objective_mismatch,
  title = 	 {Objective Mismatch in Model-based Reinforcement Learning},
  author =       {Lambert, Nathan and Amos, Brandon and Yadan, Omry and Calandra, Roberto},
  booktitle = 	 {Proceedings of the 2nd Conference on Learning for Dynamics and Control},
  pages = 	 {761--770},
  year = 	 {2020},
  editor = 	 {Bayen, Alexandre M. and Jadbabaie, Ali and Pappas, George and Parrilo, Pablo A. and Recht, Benjamin and Tomlin, Claire and Zeilinger, Melanie},
  volume = 	 {120},
  series = 	 {Proceedings of Machine Learning Research},
  month = 	 {10--11 Jun},
  publisher =    {PMLR},
  url = 	 {https://proceedings.mlr.press/v120/lambert20a.html}
}

@InProceedings{value_aware,
  title = 	 {{Value-Aware Loss Function for Model-based Reinforcement Learning}},
  author = 	 {Farahmand, Amir-Massoud and Barreto, Andre and Nikovski, Daniel},
  booktitle = 	 {Proceedings of the 20th International Conference on Artificial Intelligence and Statistics},
  pages = 	 {1486--1494},
  year = 	 {2017},
  editor = 	 {Singh, Aarti and Zhu, Jerry},
  volume = 	 {54},
  series = 	 {Proceedings of Machine Learning Research},
  month = 	 {20--22 Apr},
  publisher =    {PMLR},
  url = 	 {https://proceedings.mlr.press/v54/farahmand17a.html}
}

@inproceedings{value_equivalence,
 author = {Grimm, Christopher and Barreto, Andre and Singh, Satinder and Silver, David},
 booktitle = {Advances in Neural Information Processing Systems},
 editor = {H. Larochelle and M. Ranzato and R. Hadsell and M.F. Balcan and H. Lin},
 pages = {5541--5552},
 publisher = {Curran Associates, Inc.},
 title = {The Value Equivalence Principle for Model-Based Reinforcement Learning},
 url = {https://proceedings.neurips.cc/paper_files/paper/2020/file/3bb585ea00014b0e3ebe4c6dd165a358-Paper.pdf},
 volume = {33},
 year = {2020}
}

@misc{mecowam,
      title={Learning 4D Geometric Priors for Inference-Efficient World Action Models}, 
      author={Jianjun Zhang and Jian Zhu and Taiyi Su and Chong Ma and Zitai Huang and Yi Xu and Hanli Wang},
      year={2026},
      eprint={2607.05468},
      archivePrefix={arXiv},
      primaryClass={cs.RO},
      url={https://arxiv.org/abs/2607.05468}, 
}

@inproceedings{vggt,
  title={VGGT: Visual Geometry Grounded Transformer},
  author={Wang, Jianyuan and Chen, Minghao and Karaev, Nikita and Vedaldi, Andrea and Rupprecht, Christian and Novotny, David},
  booktitle={Proceedings of the IEEE/CVF Conference on Computer Vision and Pattern Recognition},
  year={2025}
}

@misc{robustwam,
  title         = {Robust-WAM: Bridging Generative Pretraining and Semantic Foresight in World-Action Models},
  author        = {Haodong Yan and Junfeng Li and Junjie He and Zhide Zhong and MingMing Yu and Wenxuan Song and Jiaguan Zhu and Yangyang Zheng and Yuqiao Du and Jiadi You and Yingjie Cai and Xu Yan and Guanyi Zhao and Bingbing Liu and Haoang Li},
  year          = {2026},
  eprint        = {2608.05903},
  archivePrefix = {arXiv},
  primaryClass  = {cs.RO},
  url           = {https://arxiv.org/abs/2608.05903},
}

@article{gigaworld,
  title={GigaWorld-Policy-0.5: A Faster and Stronger WAM Empowered by AutoResearch},
  author={Team, GigaWorld and Ye, Angen and Ma, Angyuan and Wang, Boyuan and Ni, Chaojun and Ye, Fangzheng and Huang, Guan and Li, Guo and Zhao, Guosheng and Yan, Haodong and others},
  journal={arXiv preprint arXiv:2607.13960},
  year={2026}
}

@article{libero,
  title={LIBERO: Benchmarking Knowledge Transfer for Lifelong Robot Learning},
  author={Liu, Bo and Zhu, Yifeng and Gao, Chongkai and Feng, Yihao and Liu, Qiang and Zhu, Yuke and Stone, Peter},
  journal={arXiv preprint arXiv:2306.03310},
  year={2023}
}

@article{robotwin2,
  title={Robotwin 2.0: A scalable data generator and benchmark with strong domain randomization for robust bimanual robotic manipulation},
  author={Chen, Tianxing and Chen, Zanxin and Chen, Baijun and Cai, Zijian and Liu, Yibin and Li, Zixuan and Liang, Qiwei and Lin, Xianliang and Ge, Yiheng and Gu, Zhenyu and others},
  journal={arXiv preprint arXiv:2506.18088},
  year={2025}
}

@article{hartigan_dip, title={The Dip Test of Unimodality}, volume={13}, ISSN={0090-5364}, url={http://dx.doi.org/10.1214/aos/1176346577}, DOI={10.1214/aos/1176346577}, number={1}, journal={The Annals of Statistics}, publisher={Institute of Mathematical Statistics}, author={Hartigan, J. A. and Hartigan, P. M.}, year={1985}, month=Mar }

@misc{adamw,
      title={Decoupled Weight Decay Regularization}, 
      author={Ilya Loshchilov and Frank Hutter},
      year={2019},
      eprint={1711.05101},
      archivePrefix={arXiv},
      primaryClass={cs.LG},
      url={https://arxiv.org/abs/1711.05101}, 
}

@InProceedings{consistency_models,
  title = {Consistency Models},
  author = {Song, Yang and Dhariwal, Prafulla and Chen, Mark and Sutskever, Ilya},
  booktitle = {Proceedings of the 40th International Conference on Machine Learning},
  pages = {32211--32252},
  year = {2023},
  editor = {Krause, Andreas and Brunskill, Emma and Cho, Kyunghyun and Engelhardt, Barbara and Sabato, Sivan and Scarlett, Jonathan},
  volume = {202},
  series = {Proceedings of Machine Learning Research},
  month = {23--29 Jul},
  publisher = {PMLR},
  url = {https://proceedings.mlr.press/v202/song23a.html}
}

@InProceedings{dmd,
  author = {Yin, Tianwei and Gharbi, Micha\"el and Zhang, Richard and Shechtman, Eli and Durand, Fr\'edo and Freeman, William T. and Park, Taesung},
  title = {One-step Diffusion with Distribution Matching Distillation},
  booktitle = {Proceedings of the IEEE/CVF Conference on Computer Vision and Pattern Recognition (CVPR)},
  month = {June},
  year = {2024},
  pages = {6613-6623}
}
